# Actuarial Applications of Natural Language Processing Using Transformers

## Case Studies for Using Text Features in an Actuarial Context

Andreas Troxler *  Jürg Schelldorfer **

v3, 25 September 2023

**Abstract**

This tutorial demonstrates workflows to incorporate text data into actuarial classification and regression tasks. The main focus is on methods employing transformer-based models. A dataset of car accident descriptions with an average length of 400 words, available in English and German, and a dataset with short property insurance claims descriptions are used to demonstrate these techniques. The case studies tackle challenges related to a multi-lingual setting and long input sequences. They also show ways to interpret model output, to assess and improve model performance, by fine-tuning the models to the domain of application or to a specific prediction task. Finally, the tutorial provides practical approaches to handle classification tasks in situations with no or only few labeled data, including but not limited to ChatGPT. The results achieved by using the language-understanding skills of off-the-shelf natural language processing (NLP) models with only minimal pre-processing and fine-tuning clearly demonstrate the power of transfer learning for practical applications.

## 1 Introduction and Overview

This data analytics tutorial has been written for the working party "Data Science" of the Swiss Association of Actuaries SAV, see

https://www.actuarialdatascience.org/

The main purpose of this tutorial is to provide an introduction to natural language processing (NLP) in an actuarial context. We start by a brief overview of NLP and traditional approaches, and then provide an introduction to NLP using transformers. We provide an overview of different workflows to incorporate NLP into actuarial contexts. We apply the concepts to two datasets which are typical in actuarial applications. The first dataset contains verbal descriptions of car accidents in both English and German language, as well as some tabular data such as the number of vehicles involved and the presence of bodily injury. The second dataset contains short descriptions of property insurance claims. The case studies in this tutorial tackle challenges arising from a multi-lingual setting, long input sequences, and also cover interpretability, which is of paramount importance in machine learning, and, in particular, in an actuarial context, where decisions need to be transparently explainable. Finally, the tutorial provides practical approaches to handle classification tasks in situations with no or only few labeled data. It also explores the use of ChatGPT for information extraction in an unsupervised setting.

* AT Analytics AG, https://www.atanalytics.ch/, andreas.troxler@atanalytics.ch
** Swiss Re, Juerg_Schelldorfer@swissre.com

This tutorial explains the concepts, case studies and results. It is accompanied by notebooks leading through the Python implementation. This tutorial, the dataset and the notebooks are available on github.[1]

## 2 What is NLP and why is it interesting and challenging?

An abundant amount of information is available in the form of text. However, language data as such is unstructured, and single words or phrases taken out of a context can be highly ambiguous. This makes it complex to exploit the data within algorithms, because computers are best suited to process structured data.

In an insurance context, text data is used in a vast number of applications and areas, such as:

- Customer interaction: Supporting the customers in their journey to purchase the right insurance product or to submit a claim.
- Underwriting: Processing of medical reports, risk survey reports, etc.
- Claims handling: Processing of first notices of loss reports to direct the information to the appropriate department or to control fully automated processing of simple and small claims; processing of loss adjuster's or expert's reports to support the setting of case reserves; detection of claims with recovery or litigation potential; etc.
- Anonymization of documents with personal information before further processing it in compliance with relevant data protection regulation.

Natural language processing (NLP) seeks to convert speech and text data into a structured data format to enable machines to

- understand the information conveyed, and use this information to support decision making (natural language understanding, NLU); or to
- formulate relevant, contextual responses (natural language generation, NLG).

Typical NLU tasks for written text data are:

- Sequence classification: Classify a sequence according to a given number of classes (e.g., classify a text document as being a medical report).
- Extractive question answering: Extract an answer to a question from a given text. The term "extractive" means that the answer is an extract of the text, as opposed to a generated new text.
- Masked language modeling: This task is designed to learn word associations from a large text corpus, an unstructured collection of possibly domain-specific text. A random fraction of words in the input text is masked, and the model is prompted to fill in the space with an appropriate word.

---

[1] https://github.com/JSchelldorfer/ActuarialDataScience/tree/master/12%20-%20NLP%20Using%20Transformers

- Causal language modeling: Fit a language model to a text corpus, by predicting the token following a sequence of tokens (also known as next-word or next-sentence prediction).
- Named entity recognition: Classify tokens according to a class, for example, identifying a token as a person, an organization or a location.
- Summarization: Summarize a document or an article into a shorter text.
- Machine translation.

In this tutorial we apply some of the above-mentioned tasks to enable the use of text data as features (or to augment other available features) for classification and regression tasks. We also explore the use of ChatGPT to extract information from texts (generative question answering).

Practical applications, in particular, in an insurance context, often come with a number of challenges:

a. The text corpus may be highly domain-specific, i.e., it may use specialized terminology.
b. Multiple languages might be present simultaneously.
c. Text sequences might be short and ambiguous. Or they might be so long that it is hard to identify the parts relevant to the task.
d. The amount of training data may be relatively small. In particular, gathering large amounts of labeled data (i.e., text sequences augmented with a target label) might be expensive.
e. It is important to understand why a model arrives at a particular prediction or classification.

This tutorial illustrates techniques to address these challenges. We use two data sets:

- Verbal descriptions of around 7'000 car accidents available in English (and translated into German), as well as some tabular data (number of vehicles involved; indicator of presence of bodily injury; etc.) which can be used as response variables to train predictive models. The length of the texts averages to around 400 words and sometimes exceeds 1'000 words.
- A dataset of ca. 6'000 property insurance claims records which include a claim amount, a very short English claim description and a hazard type with 9 different levels.

Appendix 13 provides more detail.

# 3 A brief overview of NLP approaches

The field of NLP has made significant progress over the past decades. An excellent introduction is provided in [Ferrario2020], which distinguishes three broad approaches to text classification:

1. Classical NLP pipelines based on bag-of-words and bag-of-part-of-speech methods to generate a numerical representation of text documents, used as input to a classifier;
2. Modern approaches using word embeddings (and some aggregation of word embeddings into sentence embeddings by a sentence encoder) to numerically represent text documents, again used as input to a classifier;
3. Contemporary approaches which apply a minimum degree of pre-processing and train recurrent neural networks directly on text documents.

For the sake of context, the following subsections briefly explain the key concepts of these approaches. For more details we refer to [Ferrario2020]. The remainder of this tutorial will focus on transformer models, introduced in the next section.

## 3.1 Classical NLP pipelines

This section describes the typical elements of a classical machine learning pipeline for text classification. The key steps involve:

1. text pre-processing;
2. computing a numerical representation of the text;
3. using the numerical representation as input feature for a classifier.

The following sections explain and discuss each of these steps.

### 3.1.1 Text pre-processing

The goal of text pre-processing is to transform the text to make it suitable for predictive modeling. Essentially, the text is broken down to tokens, i.e., elements of a vocabulary. This allows encoding the text by integers – the identifiers of the vocabulary elements. The required functionality is provided by libraries such as spaCy[2], NLTK[3], etc.

Figure 1 shows a fictional example.

| | | | | | | |
|---|---|---|---|---|---|---|
| raw text | `The crach occurred at an urban four-way` ***`INTERSECTION`*** `.` | | | | | |
| preprocessed text | `crach` | `occur` | `urban` | `four` | `way` | `intersect` |
| Token IDs | `[UNK]` | `7261` | `1002` | `214` | `766` | `17232` |

*Figure 1: Fictional example of text pre-processing. The text is converted to lowercase, stemmed ("occurred" becomes "occur"; "intersection" becomes "intersect") and split into tokens. Stopwords such as "the", "at" and "an" are removed. The misspelled word "crach" is not in the dictionary and is represented by a special token. In this example, punctuation is suppressed. The emphasis on the word "intersection" is lost, because formatting is suppressed and all words are converted to lowercase.*

The following paragraphs briefly explain typical pre-processing steps and some of the issues that may arise. Depending on the application, not all steps are necessarily required.

#### *Import of the raw text and formatting*

There is no such thing as clean data. Issues may arise around encoding, formatting, removal of HTML tags from web documents, etc.

#### *Conversion to lowercase*

This is a common procedure in NLP text processing, as it generates a harmonized string out of uppercase and lowercase variants of the same word. However, for some languages, the case can convey meaning (e.g., names of named entities; German nouns; etc.), so that this step could result in loss of information.

---

[2] https://spacy.io/
[3] https://www.nltk.org/

### *Tokenization*

This step splits the text into items of a vocabulary. The vocabulary is either generated based on rules from the present text data, or it is pre-defined from a large corpus of texts. In the latter case, if the present text contains words which are not present in the vocabulary, they can be mapped to a special token. This might result in a significant loss of information, for example, when the present text contains domain-specific words that are significant for the NLP task.

The following broad approaches are available:

- Word-based tokenization splits text on spaces or punctuation. Since each token represents an entire word, the information content per token is high. Limitations include: similar and closely related words (e.g., "cat" and "cats") are represented by different tokens; the vocabulary will be very large; words missing in the vocabulary will be mapped to a special token, causing a loss of information.
- Character-based tokenization splits text into individual characters. Compared to word-based tokenization, this leads to a smaller vocabulary, and out-of-vocabulary tokens will be less frequent. On the other hand, the information content per token is much smaller (unless for ideogram-based languages), and the token sequences to be processed by the model are much longer, thus, limiting the maximum allowable text length.
- Subword-based tokenization combines the advantages of the previous approaches by retaining frequent words and splitting rarer words into subwords (eg., "cats" into "cat" +"s", prefixes and suffixes). Compared to word-based tokenization, the vocabulary size is much smaller, as information is shared across different words. Many algorithms are available, such as WordPiece, Unigram and Byte-Pair Encoding.

Tokenization rules differ in many aspects, for example, in how they treat punctuation (e.g., suppression of punctuation, punctuation as separate tokens, etc.), how they treat prefixes and suffixes, or whether they apply a minimum token length. The importance of these choices depends on the language. For instance, interpunctuation rules differ between French ("n'est-ce pas") and English ("we don't"), etc.

### *Removal of stopwords*

Stopwords are frequent words with little semantic meaning. Examples might be "the" and "to" in English. Removing these words reduces the dimensionality of the vocabulary and might improve the predictive power of the NLP model. Stopwords are based on a corpus of text in a given language.

### *Part-of-speech-tagging*

Another step is tagging each token with part-of-speech (POS) attributes, such as noun, verb, adjective, etc. Models to achieve this are pre-trained on text corpora. As such, they are language-specific.

### *Stemming or lemmatization*

Stemming and lemmatization are two procedures which reduce inflectional forms to their stem by truncating pre- and suffixes, conjugations and declinations. The goal of this step is to reduce vocabulary size. It is language-specific. Stemming algorithms strip affixes off from words to reduce them to their root forms, which need not be words in the given language. For example, the English word "moving" would be reduced to "mov". Lemmatization is a more complex approach to the problem of determining a stem of a word. This process involves first determining the part-of-speech of a word, and applying different normalization rules for each part-of-speech. Often one uses a pre-trained library for this.

### 3.1.2 Computing numerical representations of the text: bag-of-word models

The previous steps have transformed a raw text into a sequence of tokens. Since the text corpus contains a defined vocabulary of unique tokens, this delivers the encoding of the raw text into a sequence of integers (representing the identifier of each token), see Figure 1.

The next step involves defining a real-valued representation of this sequence, which can be used as input feature to a classifier. One approach is the so-called bag-of-word model.

***Bag-of-word-model***

The bag-of word model maps each token sequence into a vector $x \in \mathbb{N}_0^V$, where $V$ is the vocabulary size, and the $i$-th element of $x$ is the number of occurrences of token $i$ in the sequence.

A drawback of this approach is that the ordering of words gets lost; this issue will be discussed later. Further, each word is treated equally important. However, certain words are much more frequent in the corpus than others (e.g., “the, “a”, “is” in English), but carry very little meaningful information about the actual contents of the text. If we were to feed the direct count data directly to a classifier those very frequent terms would shadow the frequencies of rarer yet more interesting terms. This drawback can be mitigated by normalization.

***Normalization***

Normalization applies weights to the token counts with diminishing importance for tokens that occur in the majority of documents. This produces a floating-point vector $\tilde{x} \in \mathbb{R}^V$. A common approach is the so-called “TF–IDF” transform. It multiplies the term frequency (TF) $x$ obtained from the bag-of-words by the “inverse document-frequency” (IDF), which is a function of the inverse of the proportion of documents which contain the term, with some additional normalization. For more details we refer to [Yates2011] and [Manning2008].

### 3.1.3 Using the bag-of-word model for text classification

Finally, the vector $\tilde{x} \in \mathbb{R}^V$ is used as input feature for a classifier, such as Naïve Bayes, tree-based methods, Support Vector Machines, neural networks, etc.

### 3.1.4 Discussion and variants

The bag-of-word method is simple to construct – all it takes is a vocabulary and a simple count.

One disadvantage is the large feature space for the classification task.

Also, the bag-of-word-mapping is not injective as the order of occurrence of the word gets lost. Therefore, the bag-of-word model cannot capture phrases and multi-word expressions, effectively disregarding any word order dependence.

Depending on the application, this loss of information may be detrimental to understanding the meaning of the text. For instance, consider the two sentences: “The driver of the minivan ignored the red traffic light and crashed into pickup truck” and “The driver of the pickup truck overlooked the red minivan and crashed into the traffic light”. They have a very different meaning, but they will be mapped to the identical feature vector. If the classification task is about predicting which driver is at fault, this leads to problems. On the other hand, if the classification task is about distinguishing car accident reports from fire claim descriptions, bag-of-words might work well.

The problem of word ordering can be mitigated by considering n-grams instead of single words (unigrams). A n-gram is a sequence of $n$ words. Instead of building a simple collection of unigrams ($n = 1$), one might prefer a collection of bigrams ($n = 2$), where occurrences of pairs of consecutive words are counted. Of course, this comes at the cost of much higher dimensionality.

## 3.2 Modern approaches based on word embeddings

### 3.2.1 Motivation and concept

As discussed in the previous section, one drawback of the bag-of-word models is the high dimensionality of the feature space – equal to the size of the vocabulary, $V$.

The technique of word embeddings, introduced by [Bengio2003], overcomes this by mapping each token in the vocabulary to a much lower-dimensional Euclidean space $\mathbb{R}^E$, $E \ll V$. This mapping is constructed in such a way that tokens with similar meaning are close in $\mathbb{R}^E$. Mathematically, the embedding can be described by a matrix $W \in \mathbb{R}^{V \times E}$; the embedding of the $i$-th token is simply the $i$-th row of $W$.

Various algorithms are available to produce embeddings. Some of the most important methods are the following:

- Word to vector algorithm (word2vec)
  [Mikolov2013a] and [Mikolov2013b] developed two algorithms based on the key idea that the similarity of words can be learned from the context in which the words are used, i.e., the surrounding words, in a given text corpus. One algorithm is based on predicting the center word from its context ("continuous-bag-of-words", "CBOW"), and the other one is based on predicting the context from the center word ("skip-gram").
- Global vector algorithm (GloVe)
  The global vector algorithm is an unsupervised word embedding methodology developed by [Pennington2014]. It is trained on the matrix of word co-occurrence counts of the entire text corpus (as opposed to one context window at a time).

For details, we refer to [Ferrario2020] and [Wüthrich2021]. Note that the approaches described in the following sections also use word embeddings.

### 3.2.2 Using word embeddings for text classification

The starting point is a tokenized text, obtained from a pre-processing process as explained in Section 3.1.1. The word embedding is applied to these tokens.

In practical applications, the volume of available text data may not be sufficiently large to train embeddings from scratch. In this case, pre-trained embeddings can be used, which are available in packages such as GloVe[4] and spaCy[5], trained on Wikipedia and other large text corpuses. The use of pre-trained embeddings speeds up the process significantly. However, it can be problematic if the text data uses a domain-specific vocabulary (or common words with a very specific meaning) which is not covered adequately by the corpus that has been used to train the embedding. Also, words with multiple meanings may be problematic, such as the English words "fire" (verb or noun); "well" (adjective, adverb, noun, verb); "will" (verb, noun); etc.

There are different ways to use word embedding in text classification applications:

- Mean pooling: Apply word embedding to all tokens of the input sequence and take the average of the embedding. This produces a vector in $\mathbb{R}^E$, which is used as input feature to a classifier. With this approach, the order of the token sequence gets lost.
- Feed all the word embeddings of the text sequence into a dense neural network with an activation function suitable for classification. To ensure that all inputs have the same length, long sentences are truncated and short sequences are padded. If pre-trained embeddings are used, the embedding layer can either be kept constant, or it can be included in the training of the neural network, in order to adapt to domain-specific meanings of words. The dimension of the input feature into the neural network is $E \times T$, where $T$ is the sequence length and $E$ the dimension of the embedding.

## 3.3 Contemporary approaches using recurrent neural networks

The approaches described in the previous sections have the drawback that they either lose the information about the order of words, or they produce high-dimensional input features for the classifier.

An alternative approach is given by recurrent neural networks (RNN), which feed the input tokens sequentially through a neural network, and also feed hidden states from one step to the next, thus keeping information on the previous steps in memory. The hidden state produced by the last step is then used as input feature to the classifier. The basic architecture is shown in Figure 2. In practice, often only one or a few layers are used.

[4] https://nlp.stanford.edu/projects/glove/
[5] https://spacy.io/models

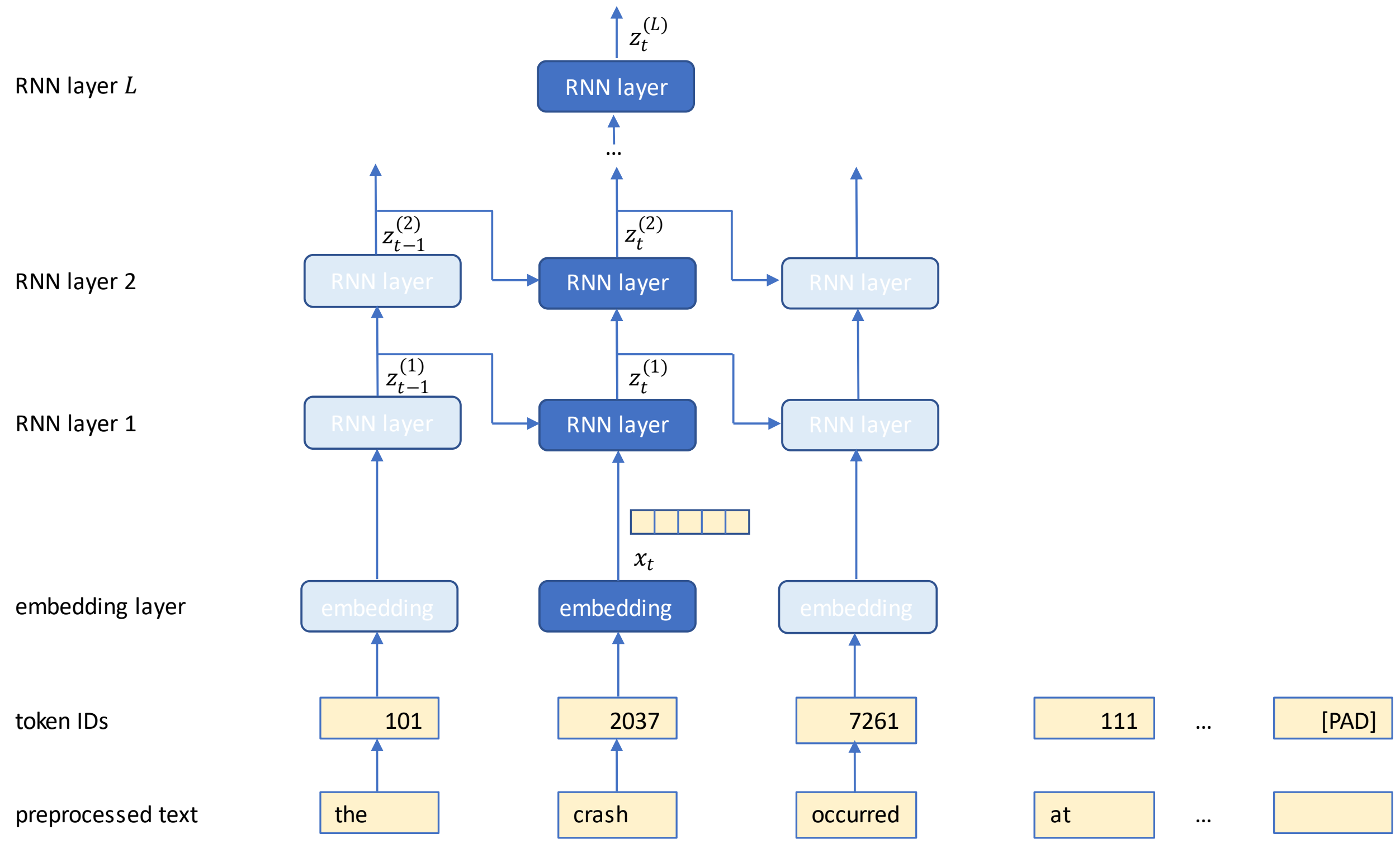

*Figure 2: Basic architecture of a recurrent neural network*

The elements of the architecture are as follows:

1. The starting point is the pre-processed text, as obtained from the workflow described in Section 3.1.1.
2. The texts are truncated to a maximum length of $T$ tokens. Shorter texts are padded with a special token.
3. For each step $t \in \{1, 2, \dots, T\}$ in the sequence, the token is passed through an embedding layer, resulting in a vector $x_t \in \mathbb{R}^E$.
4. This input is fed into one or more layers of the RNN.
   a. The first layer is fed by $x_t \in \mathbb{R}^E$ and $z_{t-1}^{(1)} \in \mathbb{R}^{H_1}$, the hidden state of the first layer from the previous step, where $H_1$ denotes the size of the first hidden layer, and with the initial value $z_0^{(1)} = 0$.
   b. The subsequent layers $l \in \{2, \dots, L\}$ are fed by $z_t^{(l-1)} \in \mathbb{R}^{H_{l-1}}$ and $z_{t-1}^{(l)} \in \mathbb{R}^{H_l}$.
5. The hidden state of the last layer of the final step, $z_T^L \in \mathbb{R}^{H_L}$ is used as input to the classifier, a dense layer with a suitable activation function for the classification task.

Note that the embedding layer and the RNN layer in item 3 and 4 are identical for all steps, i.e., they share the weights.

Different architectures are available for the RNN layers:

- Plain-vanilla dense layers. If these networks are trained with backward propagation trough time (BPTT), the necessary computation of gradients is recursive and often leads to either exploding or vanishing gradients with respect to time [Bengio1994]. This leads to numerical instability or inability to learn long-distant associations [Pascanu2013].
- Long-short-memory (LSTM) units [Hochreiter1997], which use in addition to the hidden state a cell state for long-term memory storage (not shown in Figure 2).

- Gated recurrent units (GRU) [Cho2014], a similar but less complex architecture.

We refer to Section 8 of [Wüthrich2021] and Section 5 of [Ferrario2020] for more details, and we recommend the tutorial [Richman2019].

For the embedding layer, different options are available:

- Train the embedding layer from scratch. This is computationally expensive and requires a large amount of data.
- Use pre-trained embeddings, and keep them fixed during the training of the RNN classifier. This reduces computation time and complexity significantly, but it can be problematic if the text data uses a domain-specific vocabulary.
- Use pre-trained embeddings, and update the weights of the embedding layer during the training process.

The basic architecture has been extended in different ways. For instance, bidirectional recurrent neural networks (BRNN) connect two hidden layers of opposite directions to the same output. Further, one-dimensional convolutions can be used to compress the input layer.

As can be seen from Figure 2, RNNs use sequential processing of the text: to produce the hidden state of the second token in the sequence, hidden states of the first token need to be calculated first. This inhibits parallel processing within training examples, which becomes critical at longer sequence lengths, as memory constraints limit batching across examples. This lack of parallelization leads to a performance bottleneck.

Further, the basic architecture described here follows the Markov property: each state is assumed to be dependent only on the previously seen state. This can lead to problems of processing sequences with long-range dependencies and is the reason why LSTMs and GRUs are useful.

# 4 NLP using Transformers

The drawbacks of RNNs are mitigated by the transformer architecture proposed by [Vaswani2017]. Since their introduction, transformer-based architectures have quickly become dominant for achieving state-of-the art results on many NLP tasks. This is the reason why the remainder of this tutorial focuses exclusively on transformer-based approaches.

The transformer reads the entire sequence of words at once ("bidirectional)", as opposed to directional models, which read the text input sequentially (left-to-right or right-to-left). A language model which is bidirectionally trained can have a deeper sense of language context and flow than single-direction language models, because a word may relate to other words earlier or later in the same sentence.

We recommend [Alammar2018] for an easy-to-follow introduction to the transformer architecture. [Tunstall2022] use a hands-on approach to teach how transformers work and how to integrate them in applications.

Figure 3 shows a high-level representation of the transformer architecture:

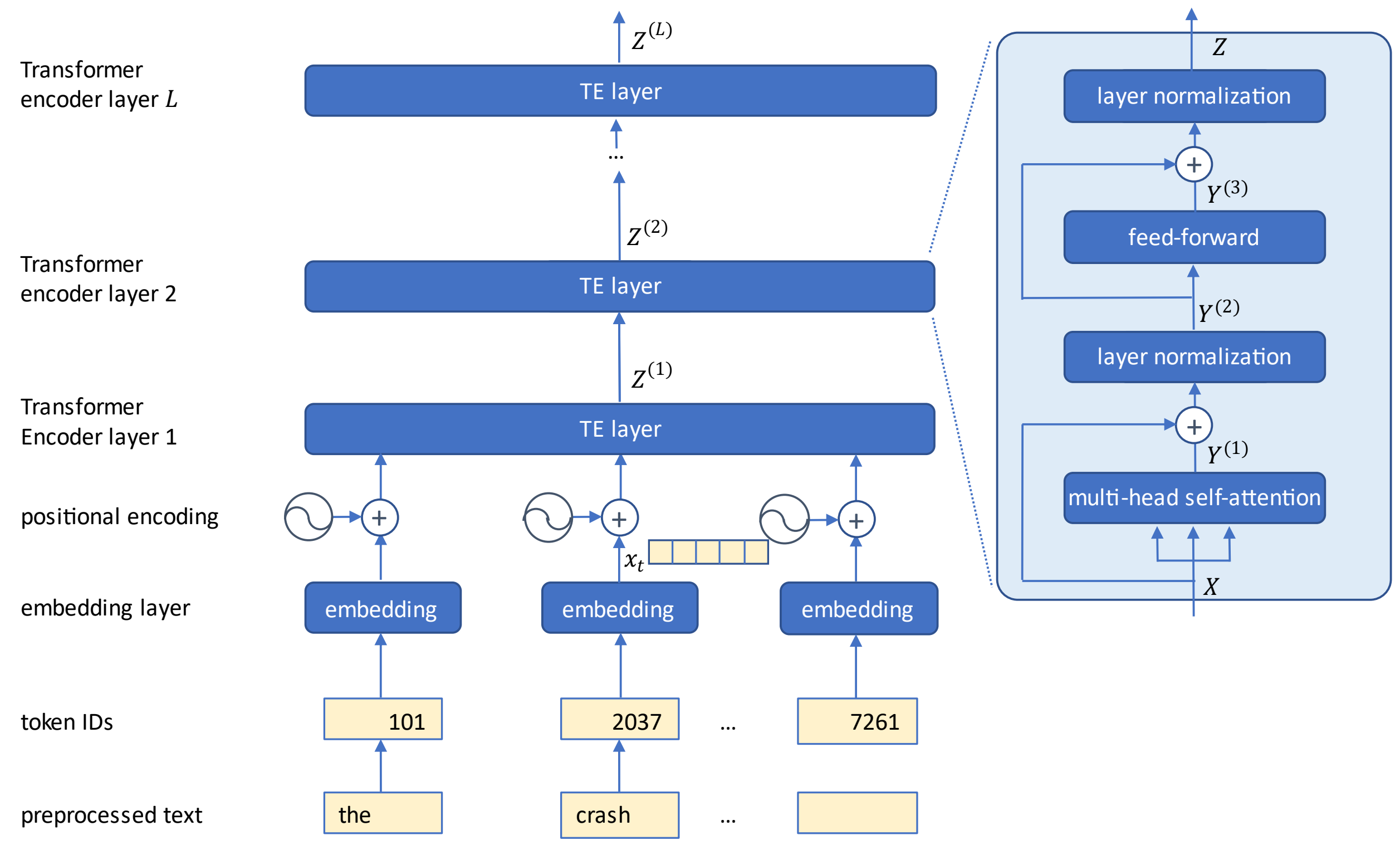

*Figure 3: Basic architecture of the transformer encoder. The architecture of a transformer encoder layer is shown in the right part of the graph.*

Comparing to Figure 2, we notice that there are no horizontal connectors and the transformer encoder layer receives the full embedded sequence at once. This reflects the non-sequential nature of the architecture. Both architectures start with an embedding layer producing the embedding $x_t \in \mathbb{R}^E$ for each token. In order to inject some information on the position of each token in the input sequence, the transformer architecture uses a "positional encoding", a vector of the same dimension $E$ as the encoding, following a fixed pattern of sine and cosine waves, which is summed to the input embedding.

The transformer encoder layer consists of a self-attention layer and a feed-forward neural network.

The motivation for using attention is to be able to model dependencies regardless of the distance in the sequence. The attention concept has been introduced by [Bahdanau2014], also referred to as "additive attention". The transformer uses the so-called dot product attention mechanism.

The self-attention layer helps the encoder to look at other words in the input sequence as it encodes a specific word. For a sequence of length $T$, it computes a matrix $A \in [0,1]^{T\times T}$ (attention matrix) with pairwise attention scores. The element $A_{i,j}$ determines how strongly the encoding of the token at position $i$ should pay attention to the token at position $j$. The rows are normalized to have unit sum. The attention matrix $A$ is calculated from the input sequence itself (hence the term self-attention). This basically gives weights according to the importance of certain words in the entire sentence.

The transformer architecture uses a multi-headed attention which runs through an attention mechanism several times in parallel. The independent attention outputs are then concatenated and linearly transformed into the dimension $\mathbb{R}^{T\times E}$. In what follows, we denote by $X \in \mathbb{R}^{T\times E}$ the layer input. For the first encoder layer, this is the embedded input sequence; for the other layers, it is the output of the previous encoder layers. The transformer encoder layer works as follows:

1. Apply the multi-head attention, with the subscript $h = 1, \ldots, H$ denoting the head:
    a. Apply three linear transformations to the input $X \in \mathbb{R}^{T\times E}$ to obtain three matrices $V_h, Q_h, K_h \in \mathbb{R}^{T\times d_K}$ (called "value", "query", "key"), where the linear transformations are represented by weight matrices $W_h^V, W_h^Q\ W_h^K \in \mathbb{R}^{E\times d_K}$ which are learned during the training process. The dimension $d_K$ is a hyper-parameter.
    b. The attention matrix is calculated from the key and value matrices:
    $$A_h = softmax\left(\frac{Q_h K_h^T}{\sqrt{d_K}}\right)$$
    The softmax function normalizes each row to unit sum, as desired. The denominator ensures well-behaved magnitude of the gradients when $d_K$ is large.
    c. The output of the self-attention is given by $Y_h = A_h V_h$ (dot product attention).
    d. Finally, the outputs $Y_h$ are concatenated and linearly transformed into $Y^{(1)} = \sum_h Y_h\, W_h^O \in \mathbb{R}^{T\times E}$, where $W_h^O \in \mathbb{R}^{H\cdot d_k \times E}$ is another weight matrix learned during the training process.
2. To speed up convergence of the training process, apply residual connection (also known as skip connection) [He2016] and layer normalization [Ba2016]:
$$Y^{(2)} = LayerNorm\left(X + Y^{(1)}\right) \in \mathbb{R}^{T\times E}$$
3. The fully connected feed-forward neural network uses two linear transformations and a ReLU activation in between, to produce $Y^{(3)} \in \mathbb{R}^{T\times E}$. The exact same feed-forward network is independently applied to each position of the sequence.
4. Apply again residual connection and layer normalization to produce the output $Z \in \mathbb{R}^{T\times E}$.

An interpretation of the attention mechanism is as follows:

- The attention matrices $A_h$ are weights that can be understood as importance weights.
- $Y_h$ is then a new representation of the values $V_h$, where the values are weighted according to their importance.

There are many different variants of the transformer architecture.

In this tutorial, we use mainly models derived from the BERT model. BERT (Bidirectional Encoder Representations from Transformers) was introduced by [Devlin2019] and developed to state-of-the-art on a large number of NLP tasks. The BERT model has over 100 million parameters, which are pre-trained on two tasks, Masked Language Modeling (MLM) and Next Sentence Prediction (NSP). This pre-training is performed on a large corpus of text data. The language mix of this text corpus determines the language(s) understood by the model.

In MLM, a random fraction of tokens of the input sequence is masked (i.e., replaced by the [MASK] token). The model then attempts to predict the original value of the masked tokens, based on the context provided by the other, non-masked, tokens in the sequence. This procedure can be performed in a fully unsupervised fashion. It is different to next word prediction which is inherently directional, thus limiting context learning.

In NSP, the model receives pairs of sentences as input and learns to predict whether the second sentence in the pair is the subsequent sentence in the original document. To help the model distinguish between the two sentences in training, the [CLS] token is inserted at the beginning of the first sentence and a [SEP] token is inserted at the end of each sentence. In addition, a sentence embedding indicating Sentence A or Sentence B is added to each token. The model output corresponding to the [CLS] token is used for the prediction. As such, this output can be seen as a representation of the full input sentence. This property will be used in Section 6.1.

The fact that the pre-training of transformer models can be performed in an unsupervised fashion on a large text corpus is very important. It enables transfer learning, i.e., transferring the language understanding skill acquired during pre-training to an applied task with a much smaller data volume. As will be shown in Section 7.1, the transformer model can be fine-tuned on the application-specific text corpus if required, at significantly reduced effort.

# 5 NLP workflows

Before moving to the applications, we describe in this section general approaches to augment tabular data for classification and regression problems with text data.

A. Use NLP techniques to extract additional features from text data (each with a specific meaning understandable by humans). Add these additional columns to the tabular data already available, and apply a supervised learning technique to the augmented tabular data. This approach is illustrated in Figure 4.
B. Use NLP techniques to encode the text sequences into a finite set of additional columns, and proceed as in Approach A. The difference to Approach A is that the meaning of the columns representing the text sequence is not defined by humans, but by the NLP encoder. This approach is illustrated in Figure 5.
C. The transformer models used in this tutorial are neural networks which encode a text sequence into a tensor. If the prediction model for the tabular data is also a neural network, the outputs of the last hidden layers of the two networks can be concatenated and the networks can be trained at the same time. This approach is illustrated in Figure 6. A variant of this approach would leave (parts of the) transformer encoder frozen during the training process.

In this tutorial, we will demonstrate Approach B in Section 6.2 and Approach C in in Section 7.2. In both cases, we will not use additional tabular data. These examples can be seen as a demonstration how to build the NLP classifier of Approach A.

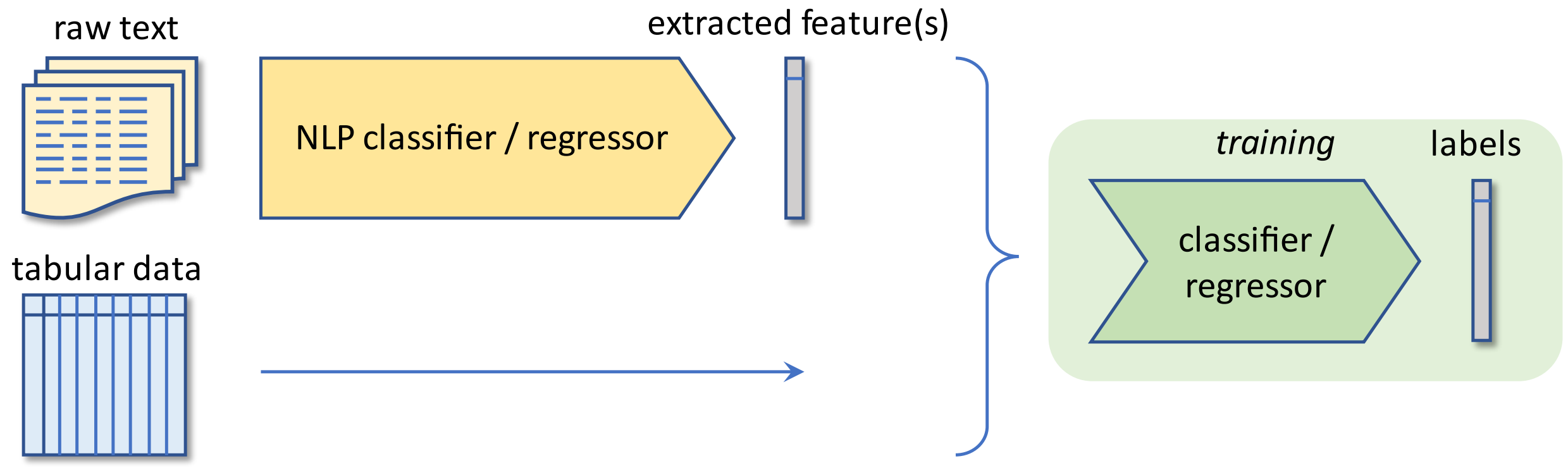

*Figure 4: NLP techniques are used to extract additional features, which are used to augment the available tabular features.*

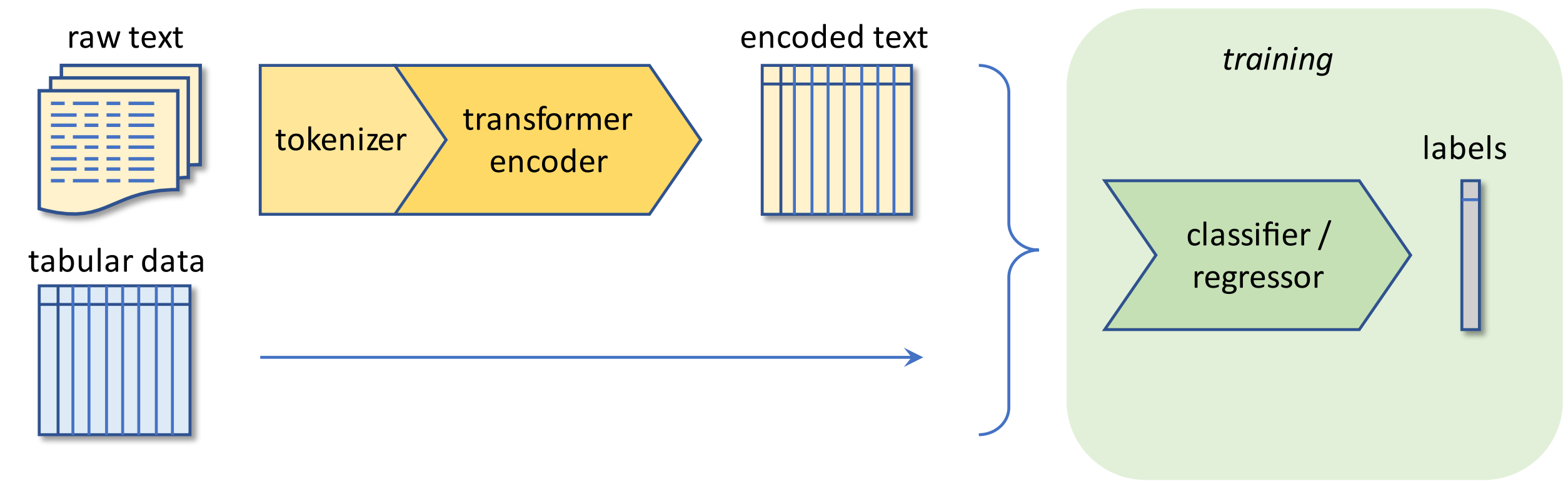

*Figure 5: An NLP encoder used to encode the text data into additional features, which are used to augment the available tabular features.*

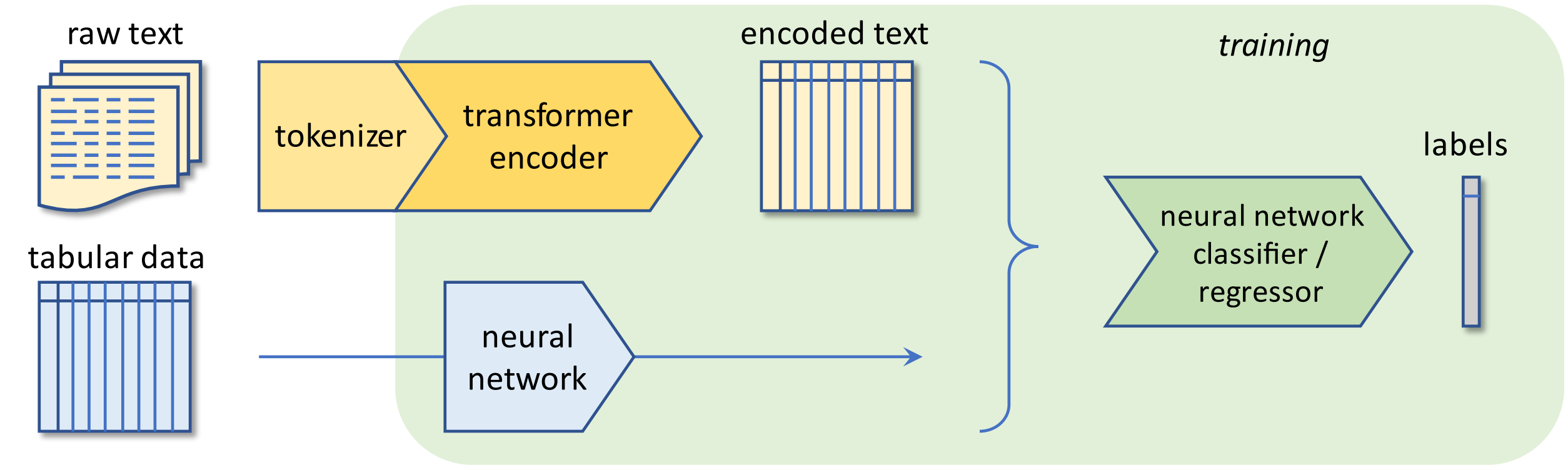

*Figure 6: The transformer encoder and the neural network to process the tabular data are connected and trained together. Parts of the transformer encoder could be frozen during the training process.*

Whilst all applications in this tutorial are classification tasks, regression tasks would be handled in an analogous fashion.

# 6 Using transformers to extract features for supervised learning tasks

This section shows how transformer models can be used to extract features from text data to feed into regression or classification tasks. This demonstrates Approach B discussed in the previous section (Figure 5). Special emphasis is put on multilingual situations. In this section, the transformer model is used off-the shelf. Fine-tuning will be discussed in the next section.

## 6.1 Basic approach

The idea is simple: the transformer encoder is used to encode the text data into a numerical representation, which is then used as input feature to a regression or classification model.

This is an example of *transfer learning:* The NLP model learns language understanding skills from a very large corpus of text data, using large-scale computing power. Both elements allow for a powerful (but relatively complex) model. This model can then be applied to situations where the availability of data or computing power would not allow for such complex models.

Throughout this tutorial, we use the transformers[6] library provided by HuggingFace[7].

In the following case study, we use the `distilbert-base-multilingual-cased` model, a multilingual BERT model, which was trained on the concatenation of Wikipedia in 104 different languages.[8] DistilBERT, introduced by [Sanh2019], is a simplification of a BERT model with fewer parameters and a faster execution time but similar language understanding capabilities as BERT. "cased" in the name of the model refers to the fact that the model differentiates words by case. The sequence length of this model is limited to a maximum of $T = 512$ tokens.

As explained in Section 4, the dimension of the transformer encoder output is $\mathbb{R}^{T\times E}$, where $T$ is the sequence length and $E$ is the output dimension of the last layer (768 for our model). We wish to represent the entire encoded sequence in one vector in $\mathbb{R}^{E}$, regardless of the sequence length. The following approaches are available and used in practice:

a) Using the first element of the encoder output, corresponding to the `[CLS]` token. As explained in Section 4, the motivation behind this approach is that pre-training of BERT on next sentence prediction tasks uses the encoder output corresponding to the `[CLS]` token as input to the prediction.
b) Mean pooling, i.e., averaging over the encodings of all sequence items (except those items which are used to pad the sequence to a minimum length).

Note that the transformer encoder and its self-attention mechanism do have a sense of word ordering. As such, mean pooling on its output does not loose information on word ordering, in contrast to mean pooling applied directly on input embeddings, mentioned in Section 3.2.2.

In the following case study, we will explore both approaches. We use a simple multinomial logistic regression classifier (see for instance Section 4.4. of [Hastie2008]) from the scikit-learn[9] library, with L2-regularization to avoid overfitting.

## 6.2 Case study 1: Use English accident reports to predict the number of vehicles involved

In the case study presented in this section, we use accident reports from the dataset described in Appendix 13.1 to predict the number of vehicles involved. In contrast to Figure 5, we do not use any other (tabular) data for this task. The column `NUMTOTV` available in the dataset is used as labels to train a classifier in a supervised learning setting.

Figure 28 in the appendix shows the distribution of the number of vehicles. Most cases involve two vehicles, and only very few cases involve more than three vehicles. No cases involve zero vehicles. We could formulate the prediction problem as regression problem, with a suitable distributional assumption. Because of the zero mass at 0 vehicles and a relatively low mass at high vehicle counts, the (zero-truncated) Poisson distribution does not appear to be a good representation of reality. Therefore, we formulate the prediction problem as multiclass classification. To avoid a heavily unbalanced classification problem, we aggregate all cases of three or more vehicles into one group. In the following, the three resulting levels are denoted by 1, 2 and 3+.

[6] https://huggingface.co/docs/transformers/index
[7] https://huggingface.co/
[8] https://github.com/google-research/bert/blob/master/multilingual.md#list-of-languages
[9] https://scikit-learn.org/stable/

We proceed as follows:

1. Train-test split: 80% of the records are used as training set and the remainder as test set.
2. Tokenize both sets using the standard tokenizer that ships with `distilbert-base-multilingual-cased`. This is a WordPiece tokenizer (see Section 3.1.1).
3. Apply the model to obtain the outputs of the final layer – the encoded sequence.
4. Condense the encoded sequence into a single vector using one of the approaches described in the previous section.
5. Use this vector as input feature for a multinomial logistic regression classifier with L2-regularization (GLM), and train the classifier to predict `NUMTOTV`.

Note that only step 5 is task-specific. For the first steps, we use the transformer model without any further training.

Regularization in step 5 is crucial to mitigate the risk of overfitting. In this setting, we have 768 features (the fixed output dimension of the NLP model) and only between 5'000 and 6'000 training samples.

Table 1 and Figure 7 show the results, evaluated on the test set. As a baseline, we also show results for a dummy classifier which always predicts the most frequent class. For a discussion and explanation of different scoring metrics see [Fissler2022].

| model | log loss | Brier loss | accuracy score |
|---|---|---|---|
| Dummy classifier | 0.961 | 0.574 | 57.2% |
| Logistic regression classifier, a) using first element of encoded sequence | 0.275 | 0.146 | 90.9% |
| Logistic regression classifier, b) using mean pooling of encoded sequence | 0.127 | 0.063 | 96.0% |

*Table 1: Scores of the different approaches, evaluated on the test set.*

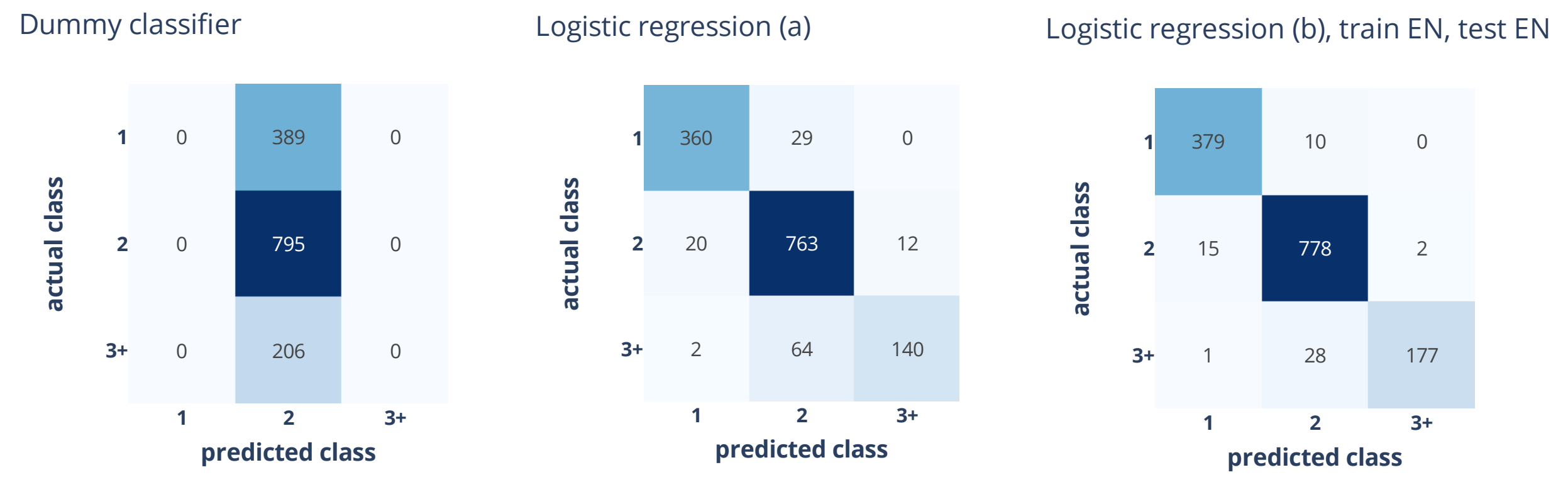

*Figure 7: Confusion matrices, evaluated on the test set. The Dummy classifier always predicts the most frequent class.*

In this case, mean pooling outperforms the use of the encoding corresponding to the first token. For this reason, we use mean pooling in what follows.

To conclude, the quite impressive accuracy of 96% clearly demonstrates the power of transfer learning. We have used the language-understanding skills of the NLP model off-the-shelf, without any fine-tuning. We have only trained a simple classifier on its output.

## 6.3 Case study 2: Cross-lingual transfer and multi-lingual training

In practice, text data might be present in different languages. This is usual in Switzerland, where the primary insurers deal with four official languages and English. As an example, claims descriptions exist in these five languages.

One option is to translate the data to a single language, for example using DeepL[10] or any other machine translation software, as a preprocessing step before applying the NLP model. A possible drawback of this approach is that machine translation might not be available in the desired quality for the languages (or, domain-specific vocabulary) present in the data set.

An alternative approach, adopted here, is to use a multilingual NLP model, such as `distilbert-base-multilingual-cased`.

Our dataset provides the accident reports in both English and German, so we can assess the performance if we train the classifier on the encoding of the German reports instead of the English reports. As Table 2 shows, similar scores are reached as with the English encodings.

| classifier trained on | test data | log loss | Brier loss | accuracy score |
|---|---|---|---|---|
| English | English | 0.127 | 0.063 | 96.0% |
| English | German | 1.083 | 0.527 | 66.0% |
| German | English | 8.052 | 1.361 | 24.3% |
| German | German | 0.120 | 0.062 | 96.0% |
| 80% English, 20% German | English | 0.136 | 0.068 | 95.7% |
| 80% English, 20% German | German | 0.160 | 0.080 | 95.2% |

*Table 2: Scores in a multi-lingual setting, using `distilbert-base-multilingual-cased`, evaluated on the test set.*

[10] https://www.deepl.com/pro-api

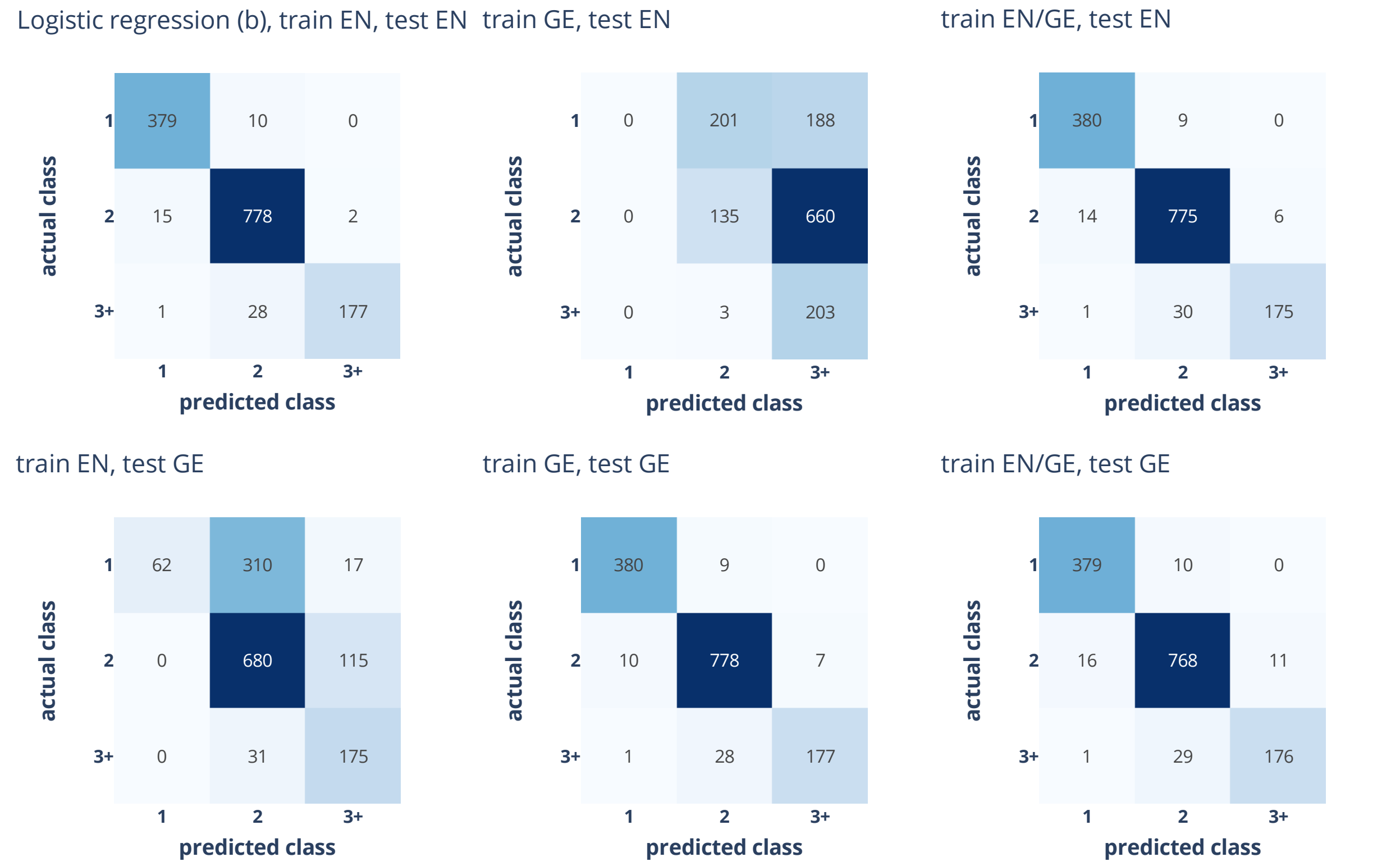

*Figure 8: Confusion matrices in a multi-lingual setting, using `distilbert-base-multilingual-cased`, evaluated on the test set.*

The next question is whether we can train the classifier on the encoded texts in one language and apply it (with no further training) to the other language. Unfortunately, this transferability is not guaranteed, and the results in Table 2 and Figure 8 show poor performance. Given the positive results of the previous experiments, we suspect that this is not caused by the NLP encoder but by the classifier, which might pre-dominantly use features of the encoded sequence which are specific to the training language.

A possible solution is to train the classifier with a training set consisting of encoded samples from both languages. In practice, this could be achieved by translating a fraction of the data and then train the model on encodings of the mixed-language data.

To simulate a situation where one language is underrepresented (as might be the case in Switzerland), we create a mixed-language dataset with 80% English and 20% German samples and train the classifier on the encoded output. This improves all scores and brigs them much closer to the mono-lingual case.

To conclude, a multi-lingual situation can be handled by a multi-lingual transformer model. For the best performance, the classifier should be trained on the encoded sequences from all languages.

# 7 Improving the model

In the case studies of the previous section the NLP model was used without any adaptation to the text data at hand.

For the task at hand, the results are already very good. However, in certain situations it might be required to further improve model performance.

In this section we will explore two approaches how to fine-tune a transformer model:

- *Domain-specific fine-tuning* involves updating the parameters of the transformer model using text data which is relevant to the domain where the model will be applied to. However, the model is not necessarily tuned for a specific downstream task of interest.
- *Task-specific fine-tuning* uses domain-specific text data and tunes the parameters of the transformer model while training it for a given downstream task of interest.

One advantage of the first approach is that it can be performed in an unsupervised fashion, i.e., it does not require labeled data. Also, the fine-tuned model can be used on different downstream tasks in the same domain.

On the other hand, task-specific fine-tuning is expected to produce better performance on the particular task which the model was tuned to, so it might be the method of choice if there is a single downstream task and if sufficient labeled data is available.

Let's explore these two fine-tuning approaches in turn.

## 7.1 Domain-specific fine-tuning

Domain-specific fine-tuning can be achieved by applying the model to a "masked language modeling" task. This involves taking a sentence, randomly masking a certain percentage of the words in the input, and then running the entire masked sentence through the model which has to predict the masked words. This self-supervised approach is an automated process to generate inputs and labels from the texts and does not require any humans labeling them in any way. Figure 9 illustrates this approach.

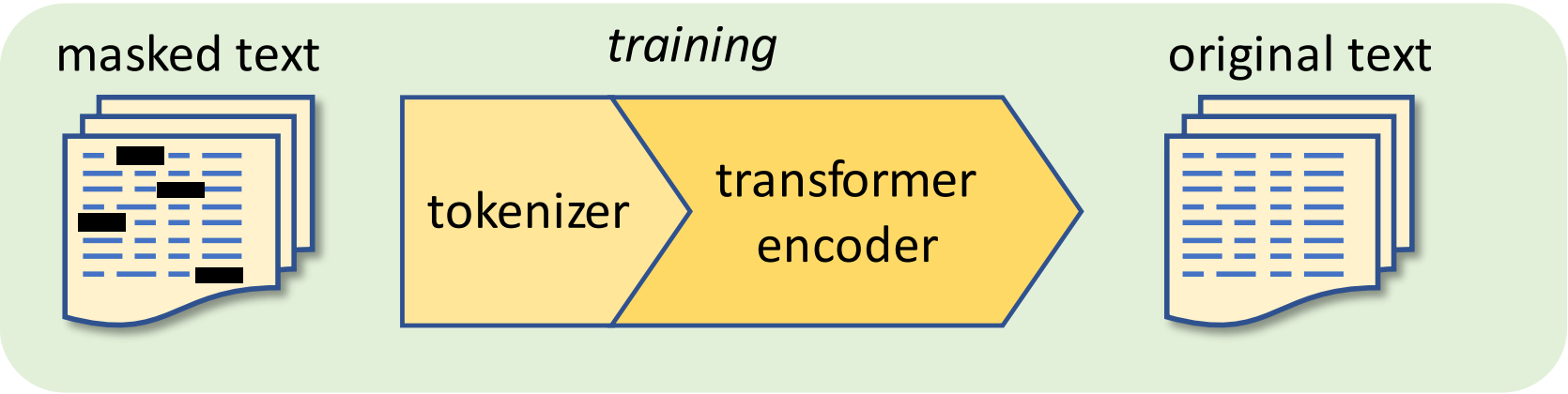

*Figure 9: Domain-specific pre-training by masked-language modeling.*

Fortunately, this process does not require much coding, as can be seen in the accompanying notebook. However, it requires significant computing resources, in particular for longer text sequences, albeit, of course, much less than training a model of comparable quality from scratch.

We apply two epochs of fine-tuning (i.e., we run the masked-language-modeling process twice through the mixed-language training set) and repeat the experiments of Section 6.3. The results are shown in Table 3 and Figure 10, evaluated on the test set.

| classifier trained on | test data | log loss | Brier loss | accuracy score |
|---|---|---|---|---|
| English | English | 0.093 | 0.044 | 97.1% |
| English | German | 0.352 | 0.207 | 85.4% |
| German | English | 2.028 | 0.631 | 65.3% |
| German | German | 0.107 | 0.053 | 96.3% |
| 80% English, 20% German | English | 0.098 | 0.047 | 97.1% |
| 80% English, 20% German | German | 0.134 | 0.063 | 96.2% |

*Table 3: Scores, evaluated on the test set, in a multi-lingual setting, using `distilbert-base-multilingual-cased` with 2 epochs of masked-language-modeling on the mixed-language training set.*

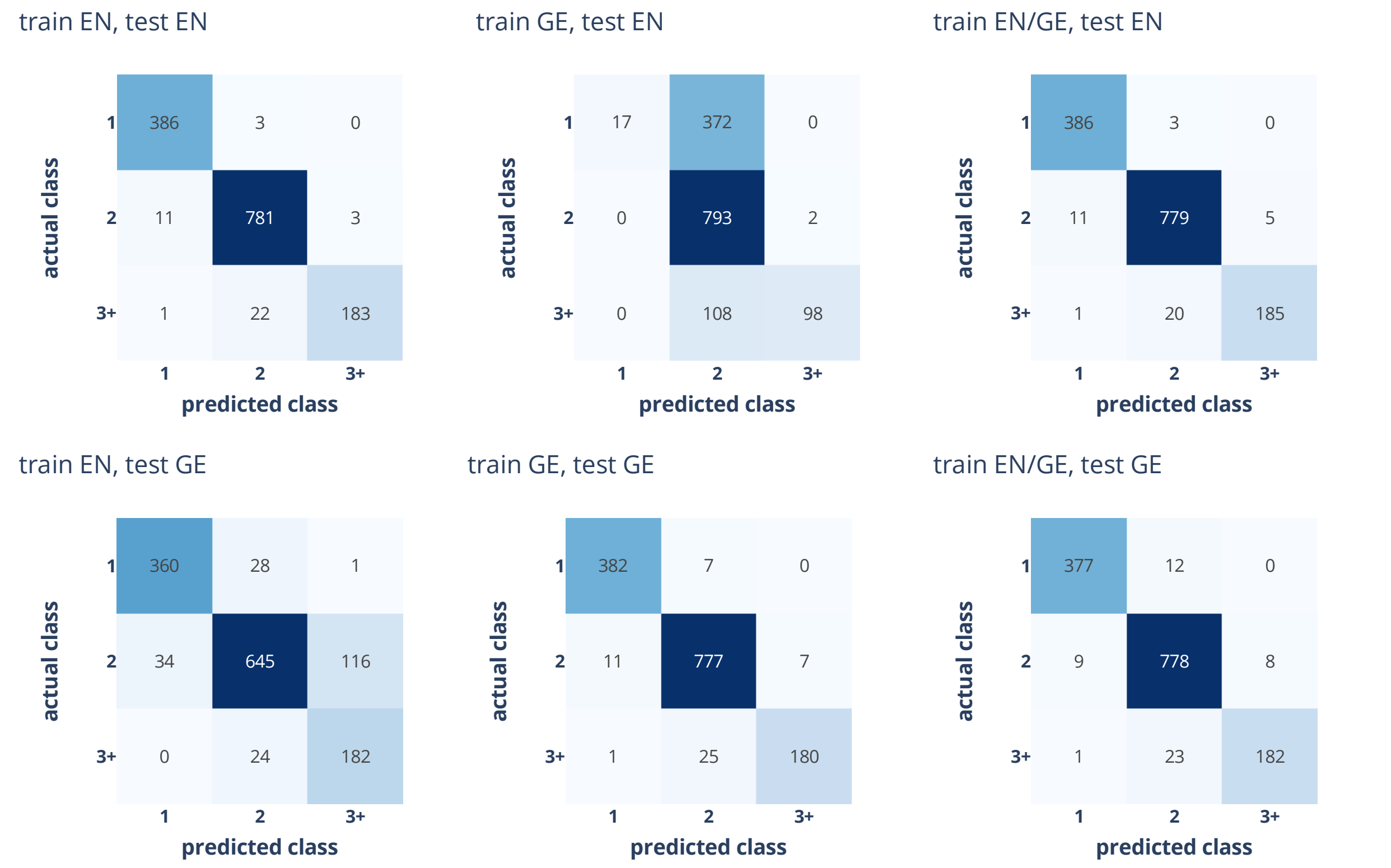

*Figure 10: Confusion matrices, evaluated on the test set, in a multi lingual setting with 2 epochs of masked-language-modeling on the mixed-language training set.*

In conclusion, by comparing to Table 2, we observe that the domain-specific fine-tuning on the mixed-language training set has improved the scores, but not to a satisfactory level for the cross-language transfer cases.

Note that since the fine-tuning was not in any way specific to the task of predicting the number of vehicles, the fine-tuned NLP model can be applied to other tasks based on this dataset (or another dataset in the same domain). We will perform this in Section 8.1.

## 7.2 Task-specific fine-tuning

An alternative to domain-specific fine-tuning is task-specific fine-tuning.

The idea is to train a transformer model directly on the task at hand, in our case a sequence classification task. This means that we don't use the encoder outputs directly, but we add a classification layer to the NLP model, so that the resulting model is a neural network classifier.

This is an example of Approach C discussed in Section 5 (illustrated in Figure 6), though with no additional tabular data.

Fortunately, this process does not require more coding than domain-specific pre-training, as can be seen in the accompanying notebook. However, it also requires significant computing resources.

We apply 2 epochs of task-specific pre-training on the original `distilbert-base-multilingual-cased` model using the English training set. The results are shown in Table 4.

| classifier trained on | test data | log loss | Brier loss | accuracy score |
|---|---|---|---|---|
| English | English | 0.028 | 0.009 | 99.6% |
| English | German | 0.072 | 0.030 | 98.3% |

*Table 4: Scores, evaluated on the test set, in a multi-lingual setting, using `distilbert-base-multilingual-cased` without fine-tuning, with 2 epochs of task-specific pre-training.*

We did not repeat all the experiments, as the results indicate that 2 epochs of task-specific pre-training on the English training set improved all scores significantly both on the English test set, and also on the German test set (despite of the model being trained on English data only).

For this case, cross-lingual transfer seems to work well. Looking at the sample in Figure 30 in the appendix, this might be due to the fact that the German translation of vehicle designations is quite close to English. For example, "V1" and "V2" are the same in both languages. In general, we would advise performing the task-specific fine-tuning on a multi-lingual training set.

## 7.3 Further improvements

One challenge that might appear in practical applications is that the length of the texts exceeds the maximum length of input sequences allowed by the model. For instance, the model used in the previous case study can handle at most 512 tokens. Only the first 512 tokens are processed, and the remainder is discarded. If the relevant part of the input is discarded, this will lead to false predictions. To address this issue, we need a way to process longer input sequences. Several approaches are available:

A. Applying the limited-length model to chunks of the text sequences.
B. Use a model which can handle more than 512 tokens. Transformer-based models are inefficient at processing long sequences due to their self-attention operation, which scales quadratically with the sequence length. To address this limitation, [Beltagy2020] introduced the so-called Longformer with an attention mechanism that scales linearly with sequence length, making it easy to process documents of thousands of tokens or longer.
C. Using methods to extract the part of the input sequence which is relevant to the task. An example is provided in Section 9.1.

In more general terms, the performance could be further improved by:

- applying hyper-parameter tuning on the training parameters (such as the learning rate);
- using alternative transformer models (such as XLM or XLM-RoBERTa in the multi-lingual case); or by
- using ensemble models (i.e., using the outputs of different NLP models and / or classifiers) to arrive at the final prediction.

Finally, dimensionality reduction techniques could be used to reduce the length of the encodings. In this case, UMAP (Uniform Manifold Approximation and Projection for Dimension Reduction, [McInnes2018]) is currently the method of choice, as it preserves both the local and global structure of embeddings quite well. In Section 9.4, we will see an application of UMAP.

# 8 Interpretability and error analysis

As seen in the previous section, predicting the number of vehicles from the case descriptions is a relatively easy task for the transformer model, even in a multi-lingual situation. Therefore, we turn to a more challenging problem, presented in the following case study.

## 8.1 Case study 3: Use English accident reports to identify bodily injury

In this case study, we use accident reports from the dataset described in Appendix 13.1 to identify cases which lead to bodily injuries. The original data set contains a tabular feature (`INJSEVA`), which indicates the most serious sustained injury in the accident. One could hope to use this as the label to train a text classification model. However, this information, taken from police accident reports and being supplementary to the case description, does not necessarily align well with the case description, as shown in Table 14 in the appendix. Therefore, we use a binary bodily injury indicator (`INJSEVB`) as target label.

We fit the following models:

1. Dummy classifier, which always predicts the most frequent class.
2. Using the approach described in Section 6.2, we fit a logistic regression classifier to the mean-pooled outputs of a `distilbert-base-multilingual-cased` encoder.
3. We repeat the same, with domain-specific fine-tuning as described in Section 7.1.
4. We perform task-specific fine-tuning, as described in Section 7.2.

Table 5 and Figure 11 show the results.

| **model** | **log loss** | **Brier loss** | **accuracy score** |
|---|---|---|---|
| Dummy classifier | 0.679 | 0.486 | 58.7% |
| Logistic regression classifier on mean-pooled output of DistilBERT | 0.400 | 0.259 | 80.1% |
| Logistic regression classifier on mean-pooled output of pre-trained model | 0.362 | 0.228 | 83.4% |
| Task-specific fine-tuning of DistilBERT classifier | 0.279 | 0.155 | 90.4% |

*Table 5: Scores of the different approaches, evaluated on the test set.*

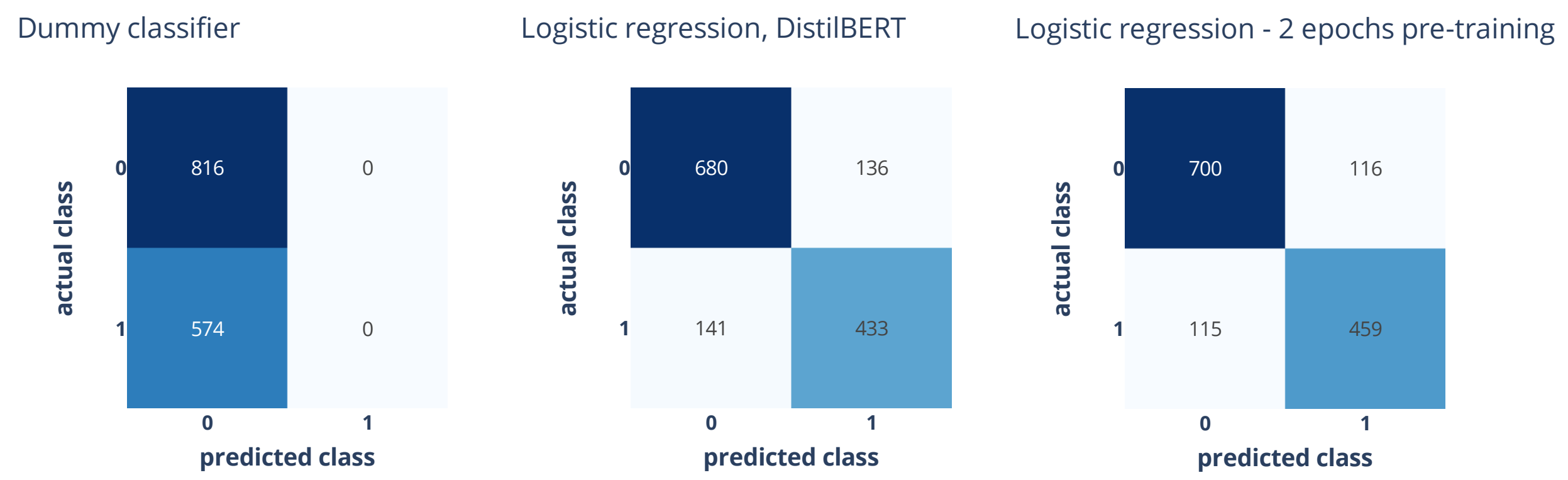

*Figure 11: Confusion matrices, evaluated on the test set. The result from task-specific fine-tuning is shown in Figure 12.*

The relative performance of the approaches aligns with the experience from the previous case studies, with task-specific fine-tuning leading to the best results.

From the confusion matrix of the neural network classifier used for task-specific fine-tuning, we observe that the false negative rate (i.e., false negatives to positives, or 1 - recall) is rather high.

The question arises how this could be improved. To answer this, we need to understand the source of the errors. This is addressed in the next paragraphs.

## 8.2 Error analysis

The first step of the error analysis is to inspect the samples producing false negative and false positive predictions. Reading every single text would be very tedious, therefore it is worthwhile focusing on those examples where the probability assigned to the false prediction was high, i.e., cases where the model was confident but wrong.

First, we look at the text lengths. For the false negative predictions, we observe that the average length of the texts exceeds 500 words. We are using a model with word-piece tokenization, so that the tokenized input sequence will be even longer on average. The transformer encoder is limited to a sequence length of maximum 512 tokens. Only the first 512 tokens will be processed, and the remainder is discarded. If only the discarded part of the text indicates the presence of a bodily injury, this will give raise to a false negative classification.

To address this issue, we adopt the approach A described in Section 7.3, as follows:

1. For the training phase, we continue using the truncated input sequences.
2. For the predictions on the test set, we split each input sequence into slightly overlapping chunks of length 512, run the prediction on each chunk, and combine the predictions by logical OR.

Fortunately, the coding effort is quite small. The results are shown in the following exhibits.

| model | log loss | Brier loss | accuracy score |
|---|---|---|---|
| Task-specific fine-tuning of DistilBERT classifier | 0.279 | 0.155 | 90.4% |
| Task-specific fine-tuning; splitting the input sequences into chunks | n/a | n/a | 92.8% |

*Table 6: Scores of the different approaches, evaluated on the test set. Since we have not implemented a logic to combine the predicted probabilities of the different chunks, the log loss and Brier loss cannot be evaluated in this case.*

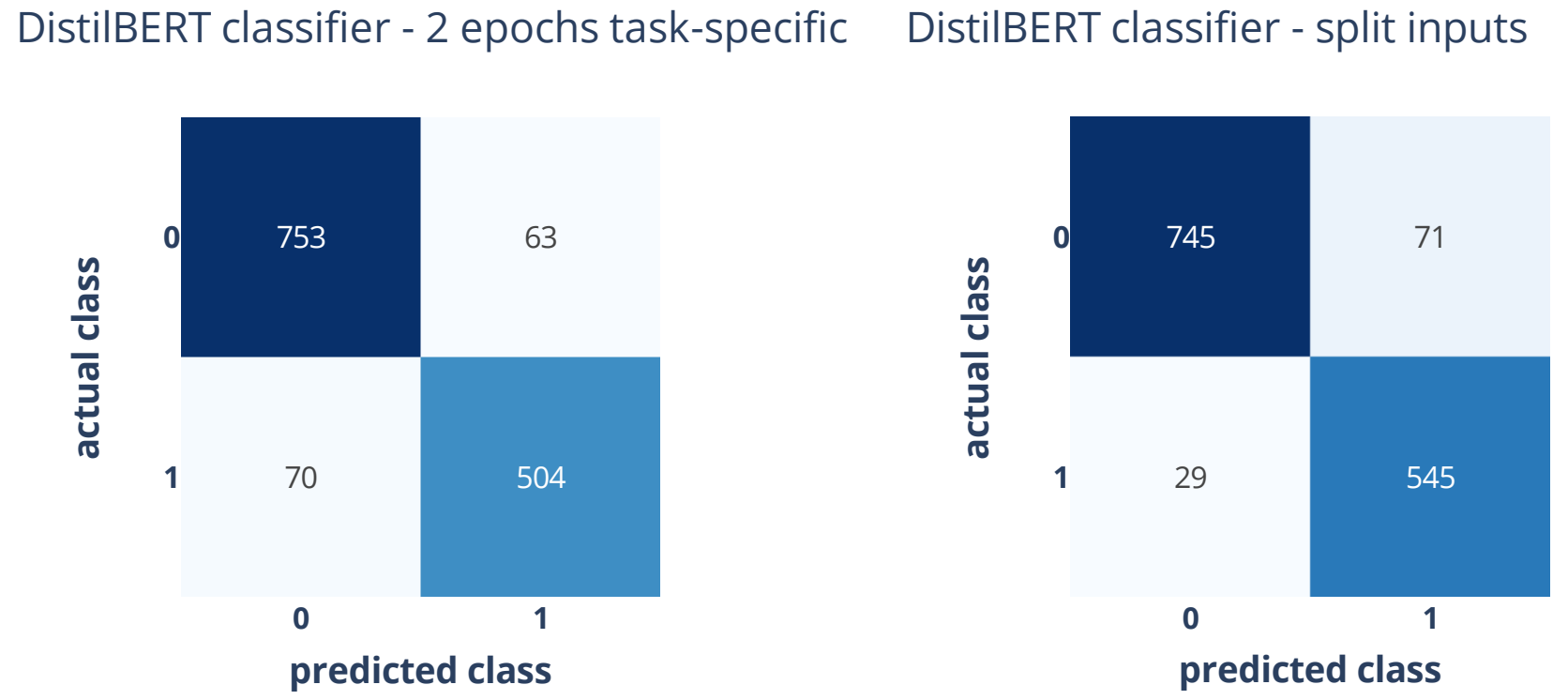

*Figure 12: Confusion matrices, evaluated on the test set.*

Indeed, splitting the input sequences has significantly reduced the number of false negatives. In turn, the number of false positives has increased slightly, due to false positives from previously discarded parts of the input texts.

To gain a better understanding of the false positives and false negatives, we would like to know why the model has arrived at a certain prediction. The next paragraph presents one way to achieve this.

## 8.3 Interpretability

Transformer models are quite complex, and therefore, interpreting model output can be difficult.

In this tutorial we briefly introduce the library Captum[11] ("comprehension" in Latin). This is an open source, extensible library for model interpretability built on PyTorch. It provides different interpretability methods, grouped into:

- Primary Attribution: Evaluates contribution of each input feature to the output of a model.
- Layer Attribution: Evaluates contribution of each neuron in a given layer to the output of the model.
- Neuron Attribution: Evaluates contribution of each input feature on the activation of a particular hidden neuron.

Here, as we primarily wish to understand which part of the text leads to a particular classification, we are going to use one of the primary attribution methods provided by Captum, namely integrated gradients.

The method of integrated gradients was introduced by [Sundararajan2017]. It represents the integral of gradients with respect to inputs along the path from a given baseline to input. Formally, it can be described as follows:

$$IG_i(x) := (x_i - x'_i) \cdot \int_0^1 \frac{\partial F\big(x' + \alpha \cdot (x - x')\big)}{\partial x_i} d\alpha, \tag{1}$$

[11] https://captum.ai/

where $F$ denotes the prediction function, $x$ the input vector of the example under consideration, $x'$ some baseline input vector, and the subscript $i$ denotes the $i$-th dimension of the respective vector. The integral can be approximated numerically. The baseline vector is a vector consisting only of padding tokens.

We use the library `transformers-interpret`[12] which provides a convenient interface to Captum, requiring only a couple of code lines.

Figure 31 to Figure 33 in Section 13.1 in the appendix show examples of word importance visualizations based in integrated gradients. We systematically evaluated these visualizations for all false positive and false negative cases, also to check that the target labels are correct. Based on such visualizations, it is possible to determine which parts of the input sequence are most important for the model to arrive at a particular prediction. This helps finding issues with the text data (e.g., ambiguous statements), erroneous labels, or shortcomings of the model.

# 9 Using transformers for unsupervised applications

The previous sections have described supervised learning approaches. This relies on the availability of sufficient labeled data. However, this is not always given in practice.

Transformer-based NLP models are also useful in unsupervised settings. The following sections introduce three unsupervised applications: extractive question answering, zero-shot classification and topic clustering.

## 9.1 Case study 4: Extractive question answering

As the name implies, the goal of extractive question answering is to extract an answer to a question from a given text. The term “extractive” means that the answer is an extract of the text, as opposed to a generated new text (which will be explored in Section 10).

For extractive question-answering, the transformer model is presented with two text sequences: the so-called context (from which the answer should be extracted), and the question. The model tries to predict the most likely start and end positions of the candidate answer(s) within the context, or the question might be unanswerable.

The question-answering is able to handle long input sequences, because it automatically splits long input sequences into chunks. Therefore, this approach is a way to extract the part of the input sequence which is relevant to the task, at the risk of losing some information. Any NLP model of choice can be applied to further process the extracted texts. Figure 13 illustrates the approach.

[12] https://github.com/cdpierse/transformers-interpret

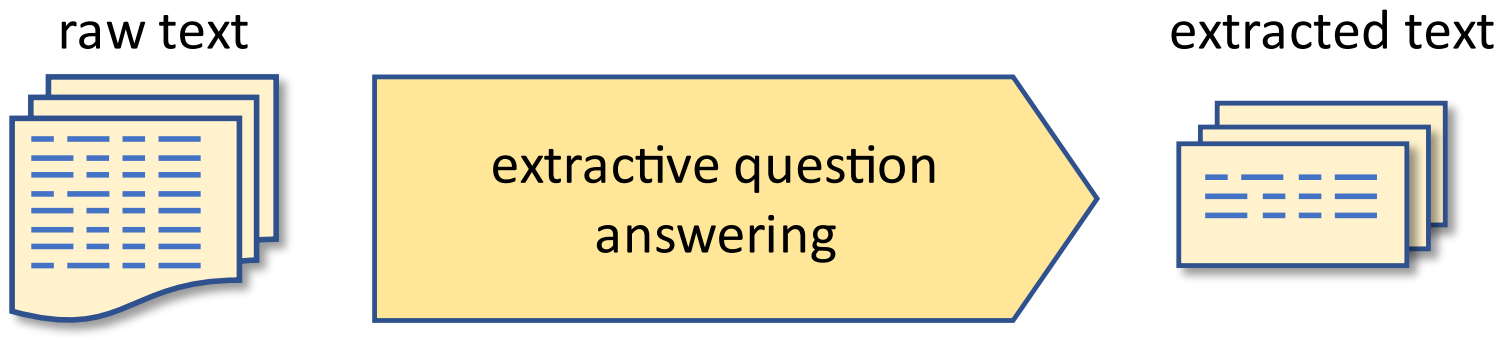

*Figure 13: Using extractive question answering to extract shorter text sequences from the raw text.*

Extractive question answering can be applied without task-specific pre-training, in a fully unsupervised fashion, thus, transferring the language understanding skills from a model that was pre-trained on a large corpus of text data. Here, we use the `deutsche-telekom/bert-multi-english-german-squad2` model.[13] The starting point to train this model was "The Stanford Question Answering Dataset" (SQUAD2.0)[14] consisting of 100k answerable and 50k unanswerable questions. These questions and answers were auto-translated into German, proofread and corrected. The `bert-base-multilingual-cased` model was fine-tuned on this question-answering task.

In the following application example, we continue the case study of Section 8.1: We apply extractive question answering to the accident reports to identify cases with bodily injury. We visit each accident report in turn (the context), and ask the model the two questions "Was someone injured?" and "Was someone transported?". Since the accident report might provide information on multiple persons, we allow a maximum of four candidate answers for each of the questions, which we concatenate into a single (much shorter) new text. For instance, the example shown in Figure 32 produces the following concatenated answers:

She was not injured. She was not injured. She was not injured in the crash. He slammed on his brakes.

her Jeep was driven from the scene. he drove off the right side of the roadway. She was transported to the hospital. the unknown vehicle passed him on the left side.

*Figure 14: Extracted answers to the questions "Was someone injured?" and "Was someone transported?" from the example shown in Figure 32 (SCASEID=2007043731967). The first four sentences are the candidate answers to the first question. The last four sentences are the candidate answers to the second question. The person who was not injured is the driver of V1; the person transported to hospital is the driver of V2.*

To assess the quality of the extracted texts, we use the approach described in Section 7.2 to train a `distilbert-base-multilingual-cased` model for the classification task. Since the text extracts are much shorter than the original texts, this training process is significantly faster. The following exhibits show the results, in comparison to those shown in Table 5 and Table 6.

| **model** | **log loss** | **Brier loss** | **accuracy score** |
|---|---|---|---|
| Dummy classifier | 0.679 | 0.486 | 58.7% |
| Logistic regression classifier on mean-pooled output of DistilBERT | 0.400 | 0.259 | 80.1% |
| Logistic regression classifier on mean-pooled output of pre-trained model | 0.362 | 0.228 | 83.4% |
| Task-specific fine-tuning; inputs sequences truncated to 512 tokens | 0.279 | 0.155 | 90.4% |
| Task-specific fine-tuning; splitting the input sequences into chunks | n/a | n/a | 92.8% |
| Task-specific fine-tuning; based on extractive question answering | 0.420 | 0.241 | 84.7% |

*Table 7: Scores obtained of the different approaches, evaluated on the test set.*

[13] https://huggingface.co/deutsche-telekom/bert-multi-english-german-squad2
[14] https://rajpurkar.github.io/SQuAD-explorer/

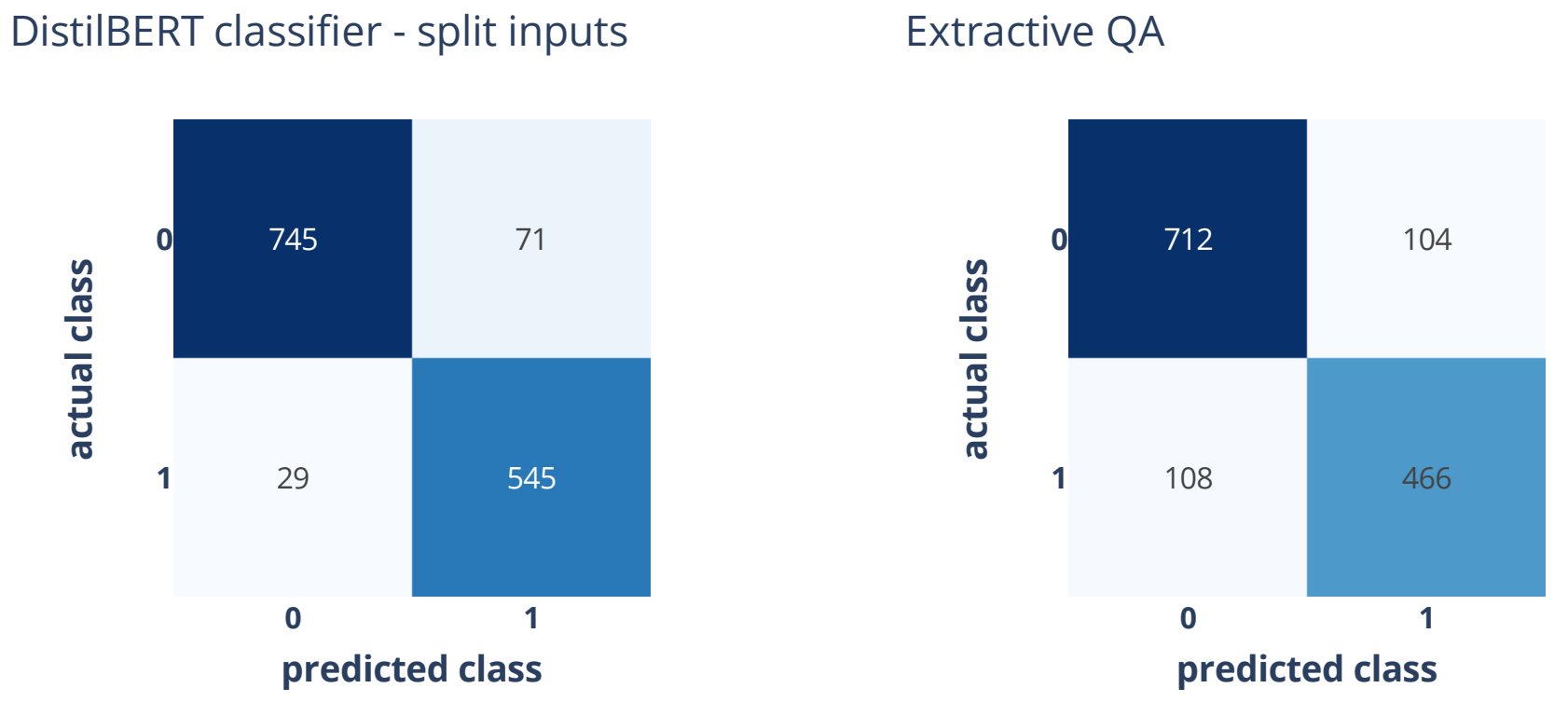

*Figure 15: Confusion matrices corresponding to the second last items of Table 7.*

From Figure 15 we see that there is a larger number of false negatives than obtained by task-specific training and evaluation on the full-length sequence. This indicates that in some cases the extractive question answering has missed out or suppressed certain relevant parts. For instance, if the original text reads "The driver was injured.", the extract "The driver" is a correct answer to the question "Was someone injured?"; however, it is too short to detect the presence of an injury from the extract.

## 9.2 Case study 5: Zero-shot classification

There are situations with no or only few labeled data. For a hands-on guide we refer to Chapter 9 of [Tunstall2022]. Here, we briefly demonstrate an approach that is suited for such situations.

Zero-shot classification is about classifying text sequences in an unsupervised way (without having training data in advance and building a model), studied by [Yin2019]. The model is presented with a text sequence and a list of expressions, and assigns a probability to each expression. Figure 16 illustrates the approach.

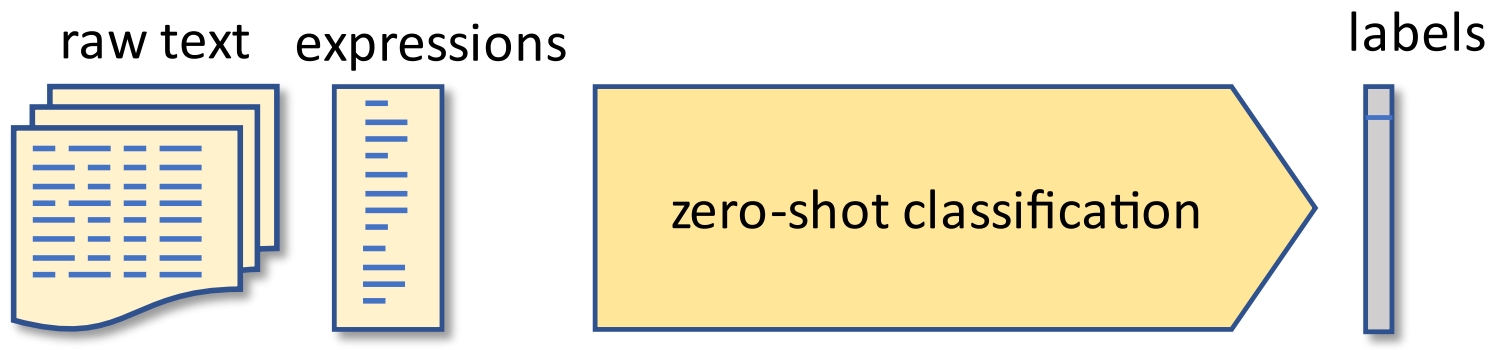

*Figure 16: Using zero-shot classification to assign probabilities to pre-defined expressions.*

We demonstrate the approach using the LGPIF dataset described in Appendix 13.2. This dataset relates to property insurance claims and consists of 6'030 records (4'991 in the training set, 1'039 in the test set). For each record, there is a short claims description in English, and a categorical variable encoding the hazard type, with the nine different levels: Fire, Lightning, Hail, Wind, WaterW (weather related water claims), WaterNW (other weather claims), Vehicle, Vandalism and Misc (any other). We define the following expressions and mapping to hazard types:

| expression | hazard type | |
|---|---|---|
| Vandalism | 0 | Vandalism |
| Theft | 0 | Vandalism |
| Fire | 1 | Fire |
| Lightning | 2 | Lightning |
| Wind | 3 | Wind |
| Hail | 4 | Hail |
| Vehicle | 5 | Vehicle |
| Water | 6 | WaterNW |
| Weather | 7 | WaterW |
| Misc | 8 | Misc |

*Table 8: Expressions and mapping to hazard types. Note that we have mapped two different expressions for the hazard type "Vandalism".*

We use the `facebook/bart-large-mnli`[15] model for zero-shot classification. The claim descriptions are the only input feature for zero-shot classification.

We apply zero-shot classification directly on the test set, because no task-specific training is required. The resulting confusion matrix is shown in the left part of Figure 17, and the scores are shown in Table 9.

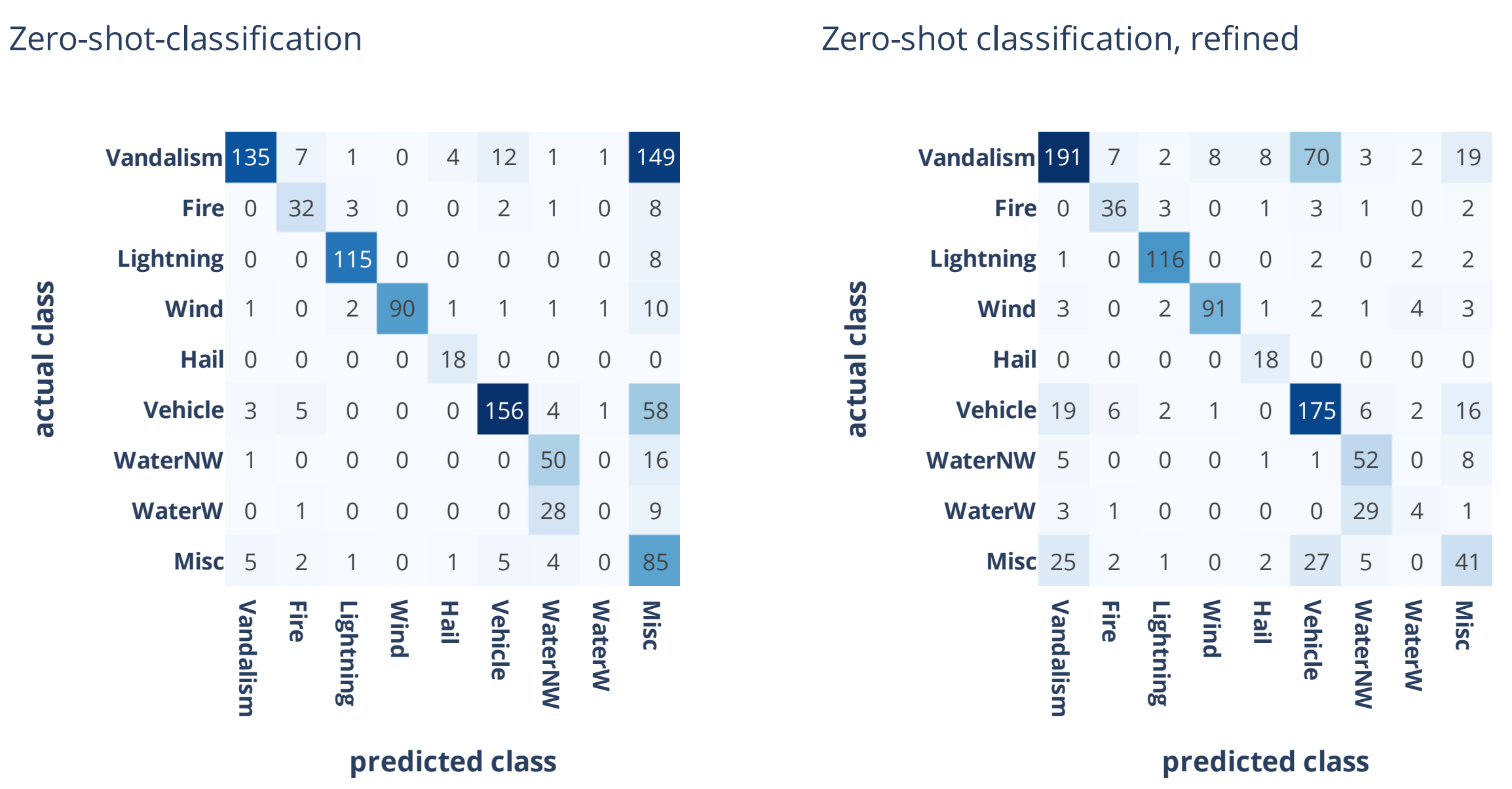

*Figure 17: Confusion matrices, evaluated on the test set.*

Apparently, the classifier struggles to correctly identify the "WaterW" cases based on the expression "Weather". Also, it seems that the expression "Misc" may not be the optimal choice, as it produces many false positives. To address this, we introduce the following heuristic: If the probability assigned to the expression "Misc" is highest but with a margin of less than 50 percentage points to the second-most likely expression, we select the latter. This increases the accuracy score from 65.5% to 69.7% and produces the confusion matrix shown in the right part of Figure 17. Compared to the 29.8% accuracy score of the dummy classifier, and considering that we have performed no training specific to the data set, this result is quite remarkable.

[15] https://huggingface.co/facebook/bart-large-mnli

Looking at false predictions in the training set, we observe the following:

a) True label "Vandalism", predicted label "Vehicle" or "Misc": Quite many descriptions contain the word "glass". For these claims, "Vandalism" appears a natural classification.
b) True label "Vehicle", predicted label "Vandalism": This group contains many descriptions like "light pole damaged", "fence damaged". Apparently, the zero-shot classifier does not realize that for these items, damage caused by a vehicle is more likely than damage caused by vandalism.
c) True label "WaterW", predicted label "WaterNW": Some of the descriptions like "frozen pipe caused water damage to indoor pool", "gutter pulled from roof ice dam", "Water damage and mold growth from storms" suggest that the candidate word "Weather" is not optimal to attract all weather-related water claims.

Based on these and similar observations, one could refine the approach by adding more candidate expressions, e.g., adding "glass" to hazard type 0 ("Vandalism"), "light pole" and "fence" to hazard type 5 ("Vehicle"), "storm" and "ice" to hazard type 7 ("WaterW"), etc.

However, we refrain from doing so because the computational effort scales with the number of samples times the number of candidate expressions. We will look at an alternative approach in the next section.

If the true labels are available (which is the case here), the supervised NLP methods described earlier in this tutorial can learn these associations automatically. To check this, we apply the methods described in Sections 6.1 and 7.2, using the `distilbert-base-uncased` model. The results are shown in Figure 18 and Table 9.

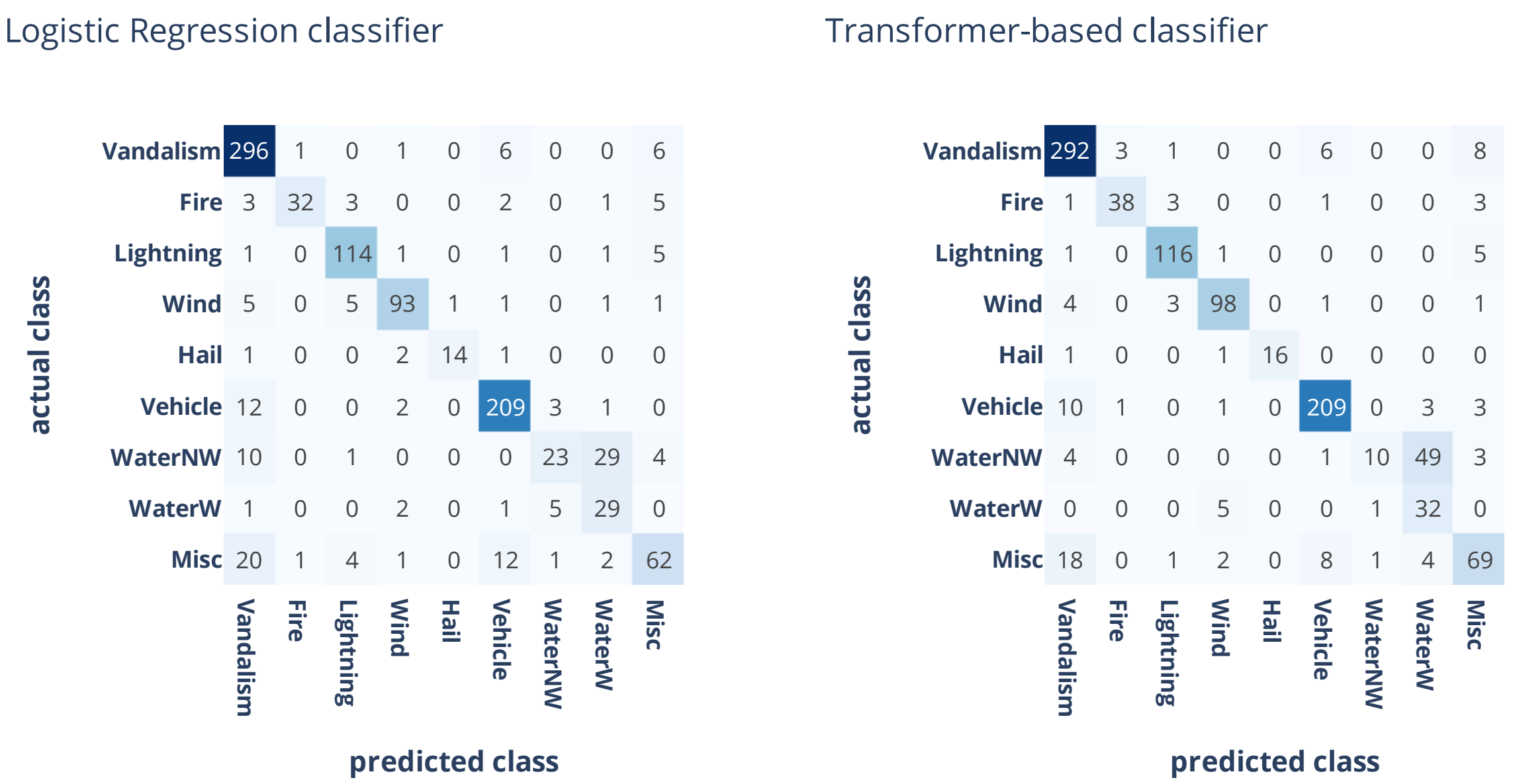

*Figure 18: Confusion matrices, evaluated on the test set.*

| model | log loss | Brier loss | accuracy score |
|---|---|---|---|
| Dummy classifier | 1.977 | 0.835 | 29.8% |
| Zero-shot classification | 1.043 | 0.463 | 65.5% |
| Zero-shot classification, with adjusted threshold for Misc | n/a | n/a | 69.7% |
| Sentence similarity | n/a | n/a | 74.5% |
| Sentence similarity, refined | 1.172 | 0.403 | 76.6% |
| Logistic regression classifier on mean-pooled output of DistilBERT | 0.531 | 0.243 | 83.9% |
| Task-specific fine-tuning of DistilBERT classifier | 0.554 | 0.233 | 84.7% |

*Table 9: Scores of the different approaches, evaluated on the test set. For the Zero-shot model with adjusted threshold, predicted probabilities are not available; therefore, the log loss and Brier loss are not shown. The sentence similarity approach will be discussed in the next section.*

As usual in machine learning, it is important to evaluate the performance on the test data, which has not been used in the training process.[16] The importance of this can be demonstrated by a simple example. For instance, the training might have a claim description "water damage to library", classified as "WaterNW". Then, the NLP classifier might learn that the association of the expressions "water" and "library" is indicative of "WaterNW" claims. However, this association has no predictive power, because water damage to a library could also be caused by a weather-related event, which needs to be classified as "WaterW".

## 9.3 Case study 6: Sentence similarity

In the previous section, we have seen that the computational effort of zero-shot classification scales with the number of samples times the number of candidate expressions, as the transformer model is applied to each pair. In this section, we explain an alternative approach with a computational effort proportional to the number of samples *plus* the number of candidate expressions. This encourages experimenting with various candidate expressions.

The idea is to encode each sample and each candidate expression separately into an embedding vector. Then, for each pair, similarity is determined by the cosine similarity score, which is the dot product of the respective embedding vectors, each normalized to unit length. For each sample, the expression with the highest similarity score is selected. Figure 19 illustrates the approach.

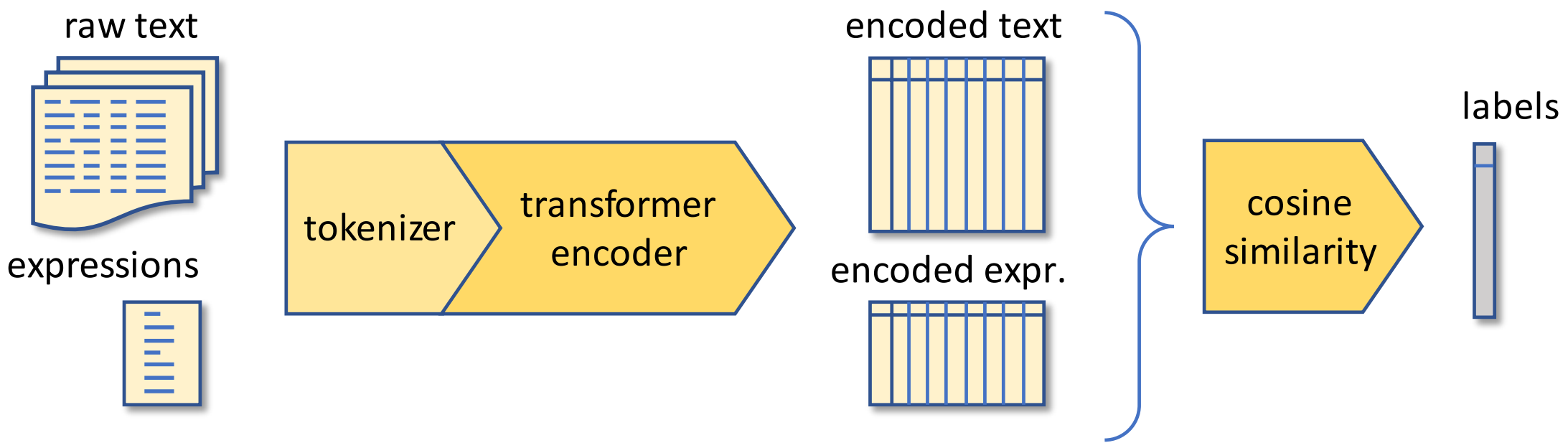

*Figure 19: Using sentence similarity to perform unsupervised classification*

[16] For this reason, our results are not comparable to those reported in [WM2021], because these were obtained by using the full data set for both training and evaluation.

We apply this approach to the test set of the LGPIF data described in Appendix 13.2, which we already used in the previous section. We define the following expressions and mapping to hazard types:

| expression | hazard type | |
|---|---|---|
| "Vandalism" | 0 | Vandalism |
| "Glass" | 0 | Vandalism |
| "Theft" | 0 | Vandalism |
| "Fire damage" | 1 | Fire |
| "Lightning damage" | 2 | Lightning |
| "Wind damage" | 3 | Wind |
| "Hail damage" | 4 | Hail |
| "Damage caused by a vehicle" | 5 | Vehicle |
| "Water damage" | 6 | WaterNW |
| "Weather damage" | 7 | WaterW |
| "Ice" | 7 | WaterW |
| "Electricity" | 8 | Misc |
| "Power surge" | 8 | Misc |

*Table 10: Expressions and mapping to hazard types.*

As you can see, we have applied some of the lessons learned from the previous experiments with the zero-shot classifier.

We use the model `sentence-transformers/all-MiniLM-L12-v2`[17], which is a BERT model that produces a sequence of real-valued vectors of length 384. During pre-training on sentence similarity tasks, mean pooling is applied to encode each the sequence into a single vector. For details, we refer to [Reimers2019].

We apply the approach directly to the test set, because no task-specific training is required. The resulting confusion matrix is shown in the left part of Figure 20, and the scores are shown in Table 9.

As a refinement, we train a sentence classifier (as in Section 7.2) on the training set, using the labels obtained by the sentence similarity approach. Although this is a supervised learning step, we are not using the original labels, therefore the overall approach is still unsupervised. The resulting confusion matrix, evaluated on the test set, is shown in the right part of Figure 20.

Compared to the zero-shot classifier, the confusion between "Vandalism" and "Vehicle" has significantly reduced, and the scores have improved. This comparison is not entirely fair, because we have not used the same candidate expressions in both cases.

The model still struggles to distinguish weather-related from non-weather-related water claims, and to identify the hazard type Misc; both difficulties are inherent in the data.

[17] https://www.sbert.net/docs/pretrained_models.html

Similarity

| actual class \ predicted class | Vandalism | Fire | Lightning | Wind | Hail | Vehicle | WaterNW | WaterW | Misc |
|---|---|---|---|---|---|---|---|---|---|
| Vandalism | 249 | 8 | 4 | 3 | 3 | 6 | 7 | 26 | 4 |
| Fire | 1 | 38 | 3 | 1 | 0 | 0 | 1 | 1 | 1 |
| Lightning | 0 | 0 | 117 | 0 | 0 | 1 | 1 | 1 | 3 |
| Wind | 3 | 0 | 2 | 90 | 2 | 0 | 0 | 10 | 0 |
| Hail | 0 | 0 | 0 | 0 | 18 | 0 | 0 | 0 | 0 |
| Vehicle | 5 | 9 | 17 | 3 | 3 | 162 | 13 | 14 | 1 |
| WaterNW | 3 | 0 | 1 | 0 | 0 | 0 | 59 | 3 | 1 |
| WaterW | 0 | 0 | 0 | 0 | 0 | 0 | 28 | 10 | 0 |
| Misc | 17 | 4 | 3 | 2 | 1 | 15 | 15 | 15 | 31 |

Simularity, refined

| actual class \ predicted class | Vandalism | Fire | Lightning | Wind | Hail | Vehicle | WaterNW | WaterW | Misc |
|---|---|---|---|---|---|---|---|---|---|
| Vandalism | 273 | 10 | 4 | 0 | 0 | 6 | 7 | 8 | 2 |
| Fire | 1 | 34 | 3 | 0 | 0 | 2 | 4 | 1 | 1 |
| Lightning | 0 | 0 | 116 | 0 | 0 | 0 | 2 | 2 | 3 |
| Wind | 3 | 0 | 3 | 88 | 2 | 1 | 0 | 10 | 0 |
| Hail | 0 | 0 | 0 | 0 | 18 | 0 | 0 | 0 | 0 |
| Vehicle | 5 | 7 | 17 | 1 | 0 | 170 | 11 | 16 | 0 |
| WaterNW | 3 | 0 | 0 | 0 | 0 | 0 | 62 | 1 | 1 |
| WaterW | 0 | 0 | 0 | 1 | 0 | 0 | 30 | 7 | 0 |
| Misc | 20 | 4 | 2 | 1 | 0 | 11 | 20 | 17 | 28 |

*Figure 20: Confusion matrices, evaluated on the test set.*

## 9.4 Case study 7: Topic clustering

In the previous sections we have seen that no prior training of the language model is required to produce a classification of reasonable quality. However, providing suitable candidate expressions is non-trivial. Ideally, we would wish for a method to extract these directly from the data.

This section describes an approach to unsupervised labeling. The idea is to encode the text samples, to create clusters of "similar" documents and to extract meaningful verbal representations of the clusters. Figure 21 illustrates the approach.

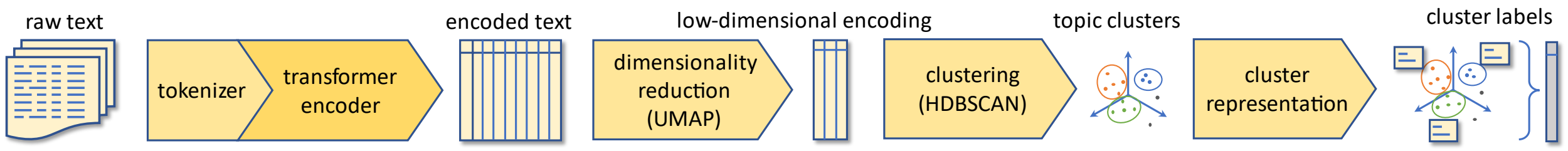

*Figure 21: Using topic clustering to generate document labels in an unsupervised setting.*

Several packages are available to perform this task, e.g., BERTopic[18], Top2Vec[19] and chat-intents[20]. These packages use similar concepts but provide different APIs, hyper-parameters, diagnostics tools, etc. Here, we use BERTopic [Grootendorst2022].

The algorithm consists of the following steps:

1. Embed documents:
    a. Encode each text sample (document) into a vector - the embedding. This can be based on a BERT model or any other document embedding technique. By default, BERTopic

[18] https://maartengr.github.io/BERTopic/index.html
[19] https://github.com/ddangelov/Top2Vec
[20] https://github.com/dborrelli/chat-intents

uses `all-MiniLM-L6-v2`, which is trained in English. In the multi-lingual case it uses `paraphrase-multilingual-MiniLM-L12-v2`.

2. Cluster documents:
    a. Reduce the dimensionality of the embeddings. This is required because the document embeddings are high-dimensional, and typically, clustering algorithms have difficulty clustering data in high dimensional space. By default, BERTopic uses UMAP (Uniform Manifold Approximation and Projection for Dimension Reduction, [McInnes2018]) as it preserves both the local and global structure of embeddings quite well.
    b. Create clusters of semantically similar documents. By default, BERTopic uses HDBSCAN [McInnes2017] as it allows to identify outliers.
3. Create topic representation:
    a. Extract and reduce topics with c-TF-IDF. This is a modification of TF-IDF (explained in Section 3.1.2), which applies TF-IDF to the concatenation of all documents within each document cluster, to obtain importance scores for the words within the cluster.
    b. Improve coherence and diversity of words with Maximal Marginal Relevance, to find the most coherent words without having too much overlap between the words themselves. This results in the removal of words that do not contribute to a topic.

We apply this algorithm to the training set the LGPIF data described in Appendix 13.2, which we already used in the previous section. The claim descriptions are the only input feature, the labels are not used. Unfortunately, reproducibility across runs is not guaranteed; therefore, we obtain different results each time.

This produces roughly 50 clusters. For each cluster, we know the samples allocated to the cluster, and we have the topic representation, as illustrated in the following figure.

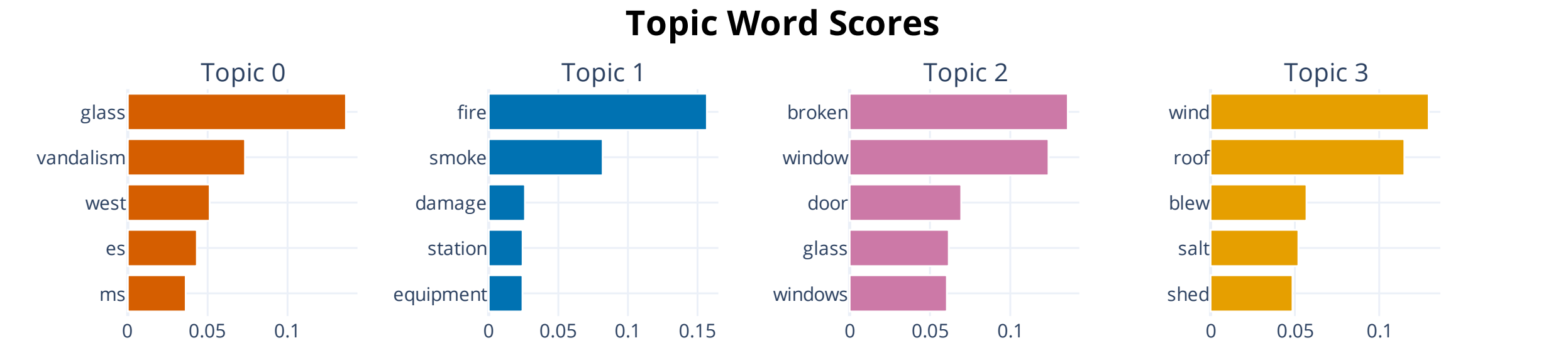

*Figure 22: Topic word scores for the first four identified topics. The topics appear in descending order of number of allocated samples. The first and third topic are mapped to "Vandalism", Topic 1 is mapped to "Fire" and Topic 3 to "Wind".*

Based on the word scores, it is straightforward to map the topics to the labels. This is typically a manual step, but performing this mapping on ca. 50 topics is much less burdensome than manually mapping thousands of samples.

In our case, we have the true labels available. We use this information to produce the mapping automatically: We map each topic cluster to the most frequent label in that cluster. A similar approach can also be used if only few labels are available. The following graph shows the relative label frequency distribution by topic.

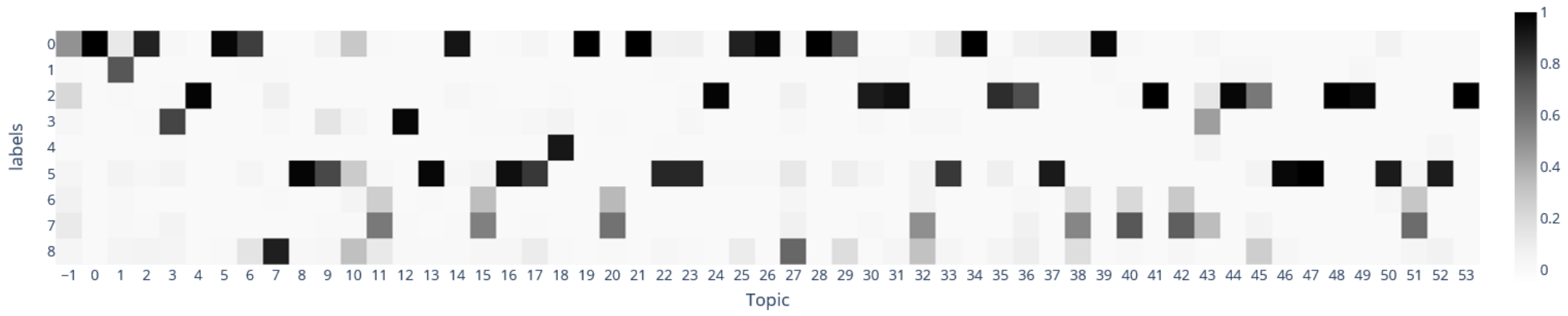

*Figure 23: The matrix shows one column per topic. The shading indicates the frequency distribution of labels within a given topic. The presence of a single dark patch indicates that almost all of the samples of the topic are associated with a single label.*

We observe that for most topics, one label is dominant, with the following exceptions:

- The classes 6 ("WaterNW") and 7 ("WaterW") are difficult to tell apart from the clusters. This finding aligns with the observations made in the previous section. All topics affected by this issue will be mapped to the more frequent class "WaterW". This results in no topic being mapped to "WaterNW".
- The first topic (-1) contains all samples which were not allocated to any of the clusters. These outliers cannot be clearly allocated to one specific label, but they will be mapped to the most frequent class 2 ("Vandalism"). This affects roughly 14% of the samples.

For all the other topics, the automatic mapping based on the true label matches the manually produced mapping.

Next, we apply the clustering and mapping to the test set. By comparing to the true labels, we obtain an accuracy score of ca. 70%. By looking at the confusion matrix in the left part of Figure 24, we observe that the class 2 ("Lightning") has many false positives. This is caused by the mapping applied to the outliers.

To mitigate this issue, we train a transformer encoder classifier on the training set using the labels obtained from the unsupervised topic modeling, but excluding the outliers. This improves the accuracy score to ca. 80%.

To conclude, unsupervised topic modeling provides an easy approach in the absence of sufficient labeled text data.

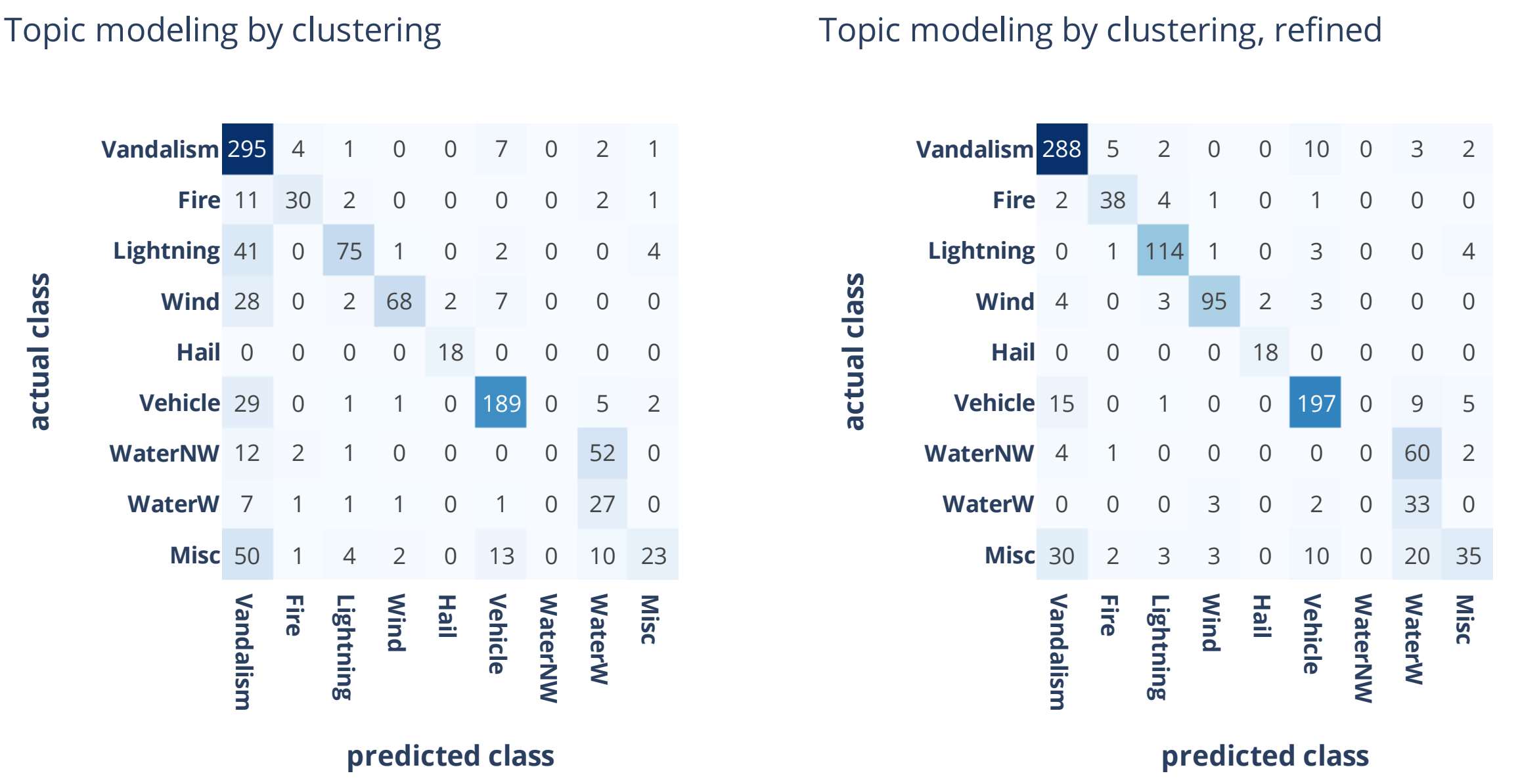

*Figure 24: Confusion matrices, comparing labels obtained from topic clustering with the true labels, evaluated on the test set. Left: Using mechanical mapping of each topic to the most frequent label. Right: Using predictions from a transformer encoder classifier trained to the labels obtained from clustering, except outliers.*

# 10 Using ChatGPT for unsupervised information extraction

The launch of ChatGPT in November 2022 has opened up completely new ways of working with unstructured data for the wider public.

ChatGPT is a chatbot developed by OpenAI[21], built upon their proprietary large language model GPT-3, [Brown2020]. GPT stands for "Generative Pre-trained Transformer". GPT is a transformer model like DistilBERT which we used in previous sections; in contrast, however, GPT uses only the transformer decoder, whilst DistilBERT uses only the transformer encoder. GPT is trained for next-word prediction on a very large multilingual corpus of natural language and program code.

In this section, we explore a possible way of using ChatGPT to extract structured information from text data, in a situation with no labels.

The idea is simple: We provide ChatGPT with a text, ask a specific question about the text, and extract the information from the answer. This approach relies on the ability of ChatGPT to understand the question and the text, and to provide an accurate answer.

While this approach is convenient, it suffers from certain limitations, including the following:

- ChatGPT may create wrong answers, due to lack of common sense, lack of detailed and up-to-date information, lack of understanding of the context, biases and prejudices, etc.
- Results obtained by ChatGPT are not reproducible across runs and model versions.

[21] https://openai.com/

- It is difficult to explain how ChatGPT arrives at the answer, and to assess its level of confidence.
- ChatGPT is a very complex model. As such, it requires significant computational resources.

These limitations should be kept in mind when evaluating this approach for a practical application.

## 10.1 Case study 8: Use ChatGPT to extract information from accident reports

In this section, we return to the accident reports from the dataset described in Appendix 13.1. As in the previous case studies 1-3, the tasks are:

- extract the number of vehicles involved;
- identify the cases which lead to bodily injuries or fatalities.

The approach is very simple: We specify a number of questions and ask ChatGPT to provide answers based on a given accident report. This is an example of generative question answering. The extractive question answering which we have employed in Section 9.1, provides in contrast extracts of the original text which are relevant to the question.

A possible prompt is:

```
Read the following text, and answer the following:
1. Was someone injured?
2. Was someone killed?
3. How many vehicles were involved?
4. Summarize your last answer by a number.
Text:
V1, a 2000 Pontiac Montana minivan, made a left turn [...]
```

*Figure 25: Example ChatGPT prompt.*

The answer might look as follows:

```
1. Yes, the driver of V2 sustained minor injuries.
2. No, no one was killed.
3. Two vehicles were involved.
4. 2
```

*Figure 26: Example ChatGPT response.*

Using standard regular expression operations, we derive from the answers to Questions 1 and 2 whether someone was injured or killed, and the number of involved vehicles from the answer to Question 4 in a straightforward manner.

Note that the response is provided in English (the language of the question), regardless of whether the accident report is presented in English or in German. This simplifies the extraction of the required information from the response.

We use the OpenAI API[22] to apply the above prompt to the first 1000 samples of the dataset (in English), using the `gpt-3.5-turbo` model with parameters `temperature=0.2` and `top_p=1`. Execution time is roughly two seconds per sample.

Figure 27 shows the confusion matrices for the two tasks, comparing to the true labels available in the dataset. Table 11 and Table 12 compare the accuracy scores to the results obtained in the previous case studies.

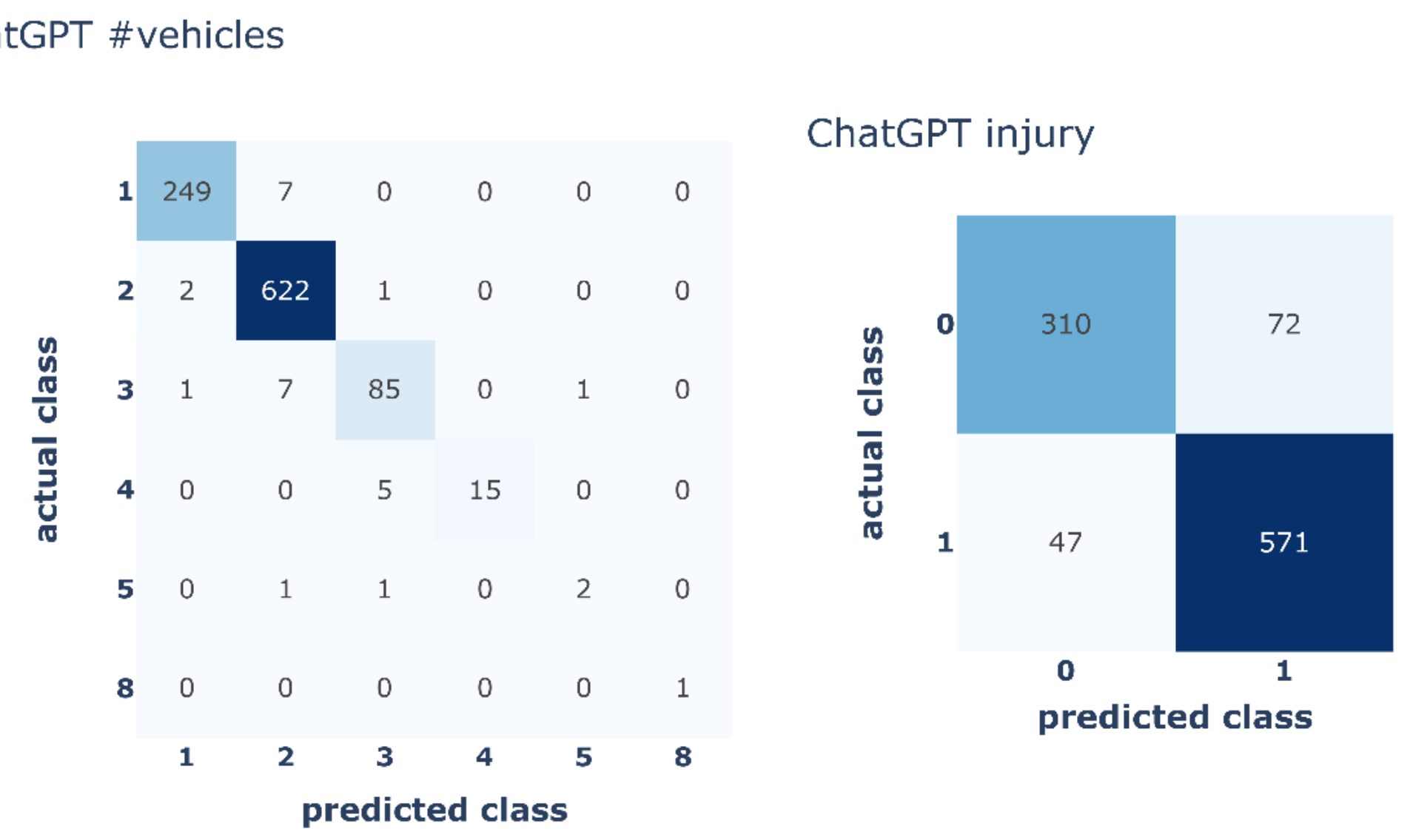

*Figure 27: Confusion matrices, comparing labels obtained from ChatGPT with the true labels, evaluated on the test first 1000 samples in the dataset. Left: Extraction of the number of vehicles involved. Right: Identification of cases leading to bodily injury or fatality.*

| **model** | **accuracy score** |
|---|---|
| Logistic regression classifier using on pooled output of DistilBERT (Table 2) | 96.0% |
| Task-specific fine-tuning of of DistilBERT classifier (Table 4) | 99.6% |
| Unsupervised information extraction using ChatGPT* | 98.1% |

*Table 11: Accuracy scores for the predicted number of vehicles involved using different models.*
** In order to make the score comparable with those shown in Table 2 and Table 4, the classes were aggregated into 1, 2, 3+. Without aggregation, the accuracy score would amount to 97.4%.*

| **model** | **accuracy score** |
|---|---|
| Logistic regression classifier using on pooled output of DistilBERT (Table 5) | 80.1% |
| Task-specific fine-tuning of of DistilBERT classifier (Table 5) | 92.8% |
| Unsupervised information extraction using ChatGPT | 88.1% |

*Table 12: Accuracy scores for the identification of cases involving bodily injury using different models.*

We observe the following:

[22] https://platform.openai.com/docs/guides/gpt

- The accuracy score is higher than with supervised training of a logistic regression classifier on the DistilBERT-encoded texts.
- The accuracy score is somewhat below the one obtained using task-specific fine-tuning of the DistilBERT model.

Note, however, that here we have not employed any task-specific fine-tuning at this stage, nor have we attempted to optimize the prompt. Possible strategies include:

- Improve clarity of the prompt. For instance, the prompt should be more specific about the evidence is allowed to identify bodily injury cases. Transportation of a person to the hospital might or might not be a suitable indication.
- To improve factuality, the instruction should ask ChatGPT to provide a response only based on the text.
- To help ChatGPT reasoning, the tasks might be split into simpler subtasks. For instance, it might be a good idea to ask ChatGPT first whether the answer to the question can be determined from the text.
- Similar to Question 4, the prompt could specify the output format.

So far we have assessed only the performance of ChatGPT in terms of the accuracy score. Ideally, we would also like to understand additional aspects:

- Calibration: We would like to evaluate the predictive uncertainty of the model. This would allow the comparison of probabilistic error measures such as log loss and Brier loss to alternative models. Simply put, we need to know how confident ChatGPT is in its responses.
- Explainability: We would like to know wy the model chooses a certain answer over the other.
- Faithfulness: Is the answer fact-based or fictitous?

Some of these aspects might be addressed by adding prompts such as "express your confidence in the previous answer as an integer between 0 and 100, 100 meaning fully confident.", "provide an explanation for your previous answer", etc.

One strategy to deal with these challenges and the limitations mentioned at the beginning of this section is a step-wise approach:

1. Use ChatGPT to produce labels, as shown before.
2. Validate the labels. If the prompt is designed so as to provide explanations, this step is much faster than manually producing labels in the first place.
3. Train a task-specific model based on the validated labels, as demonstrated in earlier sections.

We did not pursue this here, and leave the optimization of the process and prompts as an exercise for the reader.

# 11 Conclusions

In this tutorial we have provided an overview of common approaches to NLP, with a focus on transformer-based models. We have shown typical workflows to incorporate text data into classification and regression tasks, which often arise in an actuarial context.

In one of these workflows, the NLP model is used to encode the text into real-valued vectors which are then used as (additional) input features of a classifier or regressor, with no training of the NLP model whatsoever. This is an example of transfer learning: The language understanding skill of the NLP model, learned from a very large volume of text data, is transferred to an application with only limited text data available. This approach was demonstrated by predicting the number of involved vehicles and the presence of bodily injury from a real dataset of car accident reports, available in English and German. Our case study found that in a multi-lingual setting, training samples should represent all languages, albeit a heavy under-representation of one language did not lead to issues.

In our case studies, model performance was improved by domain-specific pre-training, which is a process to refine the NLP model using the available text data in an unsupervised way. We further significantly improved the model by task-specific fine-tuning and by adjusting the model to handle long input texts for prediction.

We have demonstrated how the integrated gradients method helps identifying those parts of the input text which led to a particular classification, to find issues with the text data, erroneous labels, or shortcomings of the model.

Further, we have demonstrated unsupervised techniques. Extractive question answering was used to shorten long input sequences by filtering the relevant parts. Zero-shot classification, sentence similarity and topic clustering were applied for a classification task in a situation with no labels available.

Finally, we have used ChatGPT for unsupervised information extraction, and outlined ways to address some of its limitations.

Overall, the results obtained in the case studies clearly demonstrate that transformer models provide a powerful tool to make text features useable for actuarial applications, with only minimal pre-processing and fine-tuning.

The results presented in this tutorial were obtained from an implementation in Python, available as a Jupyter notebook on github.

## Acknowledgements

The authors are very grateful to Mario Wüthrich, Christian Lorentzen and Michael Mayer for their comprehensive reviews and their innumerable inputs which led to substantial improvements of this work.

# 13 Appendix – data sets

## 13.1 NHTSA accident data

In the United States, the National Highway Traffic Safety Administration (NHTSA) is authorized by Congress to collect information on motor vehicle crashes to aid in the development, implementation and evaluation of motor vehicle and highway safety countermeasures. From 2005 to 2007, the NHTSA conducted the National Motor Vehicle Crash Causation Study (NMVCCS). This study covers a total of 6'949 cases. For each case, a text document is available, which describes the accident situation, road and weather conditions, vehicles, drivers and passengers involved, preconditions, health status, injuries of persons involved, etc. The level of detail and length of these texts varies and averages to about 400 words. In addition, tabular data is available, which encodes some of the information described before and additional information.

The NMVCCS data base consists of several tables which are linked by the case identifier (and vehicle or passenger identifier, respectively, where relevant). For simplicity, we have extracted parts of the information contained in the NMVCCS data base into a single data frame. Moreover, in order to simulate a multi-lingual environment, we have translated the accident descriptions to German using the DeepL Python API[23], without any postprocessing. Table 13 lists the columns.

| column name | type | meaning |
|---|---|---|
| SCASEID | integer | unique case identifier |
| NUMTOTV | integer | number of vehicles involved |
| WHEATHER1 … WHEATHER8 | integer | indicators of certain weather conditions |
| INJSEVA | integer | classification of most severe injury sustained |
| INJSEVB | integer | indicator of bodily injury |
| SUMMARY_EN | character string | original case description in English |
| SUMMARY_DE | character string | German translation of case description |

*Table 13: Summary of dataset columns.*

This combination of text data and tabular data is ideal for this tutorial, because it allows us the use of supervised learning techniques.

Figure 28 shows the histogram of NUMTOTV. All cases involve at least one vehicle. Most cases involve two vehicles, and only very few accidents involve more than three vehicles.

[23] https://www.deepl.com/pro-api

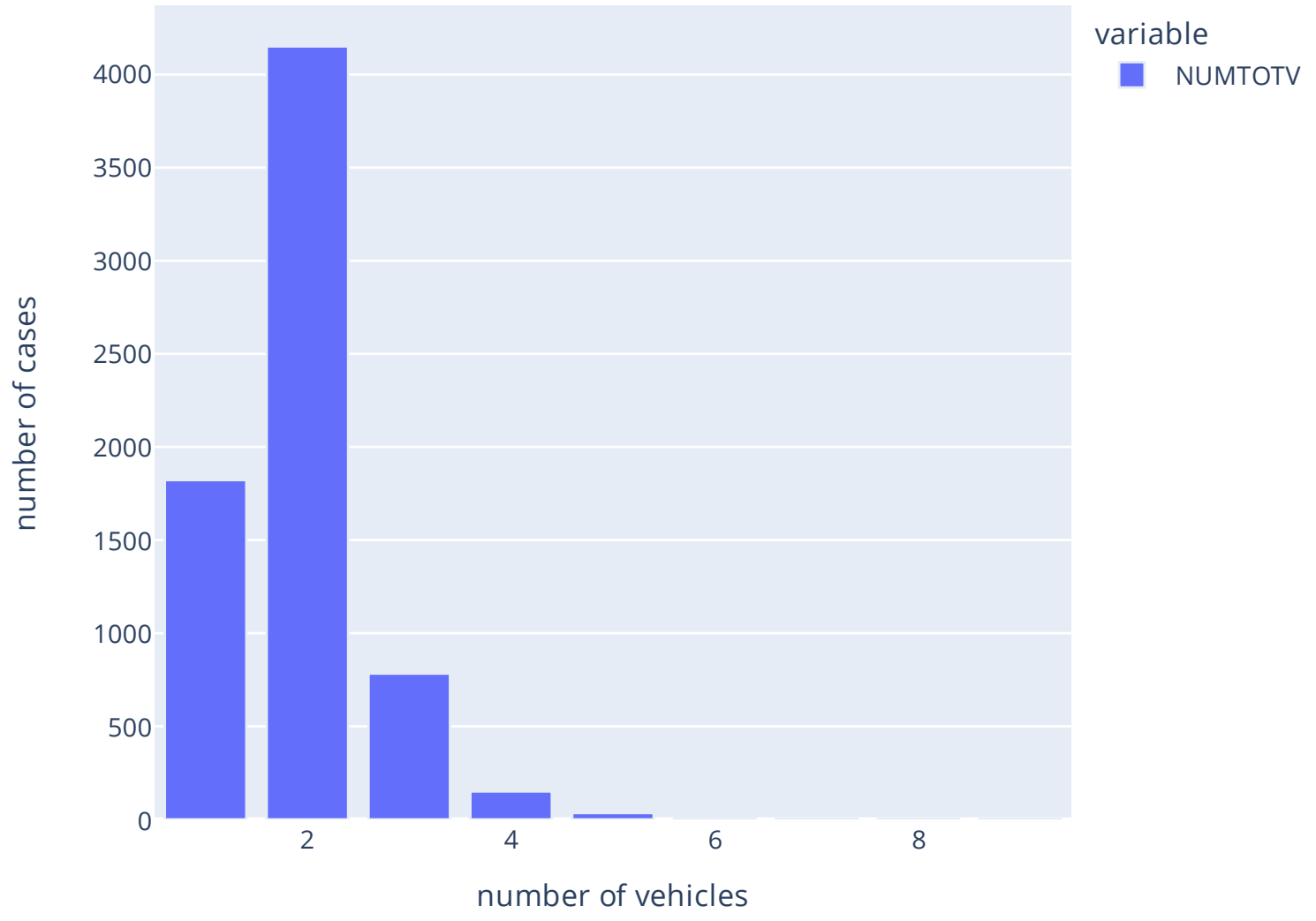

*Figure 28: Distribution of NUMTOTV, the number of vehicles involved.*

INJSEVA indicates the most serious sustained injury in the accident. For instance, if one person was not injured, and another person suffered a non-incapacitating injury, injury class 2 was assigned to the case. This information has been extracted by the NHTSA from police accident reports, if available. Unfortunately, this information does not necessarily align with the case description: There are many cases for which the case description indicates the presence of an injury, but INJSEVA does not, and vice versa. For this reason, we created manually an additional column INJSEVB based on the case description, to indicate the presence of a (possible) bodily injury. The table below shows the distribution of number of cases by the two variables.

| INJSEVA | description | INJSEVB=0 | INJSEVB=1 | total |
|---|---|---|---|---|
| 0 | O - No injury | 1’458 | 96 | 1’554 |
| 1 | C - Possible injury | 1’112 | 1’298 | 2’410 |
| 2 | B - Non-incapacitating injury | 729 | 945 | 1’674 |
| 3 | A - Incapacitating injury | 304 | 373 | 677 |
| 4 | K - Killed | 5 | 114 | 119 |
| 5 | U - Injury, severity unknown | 44 | 122 | 166 |
| 6 | Died prior to crash | 0 | 0 | 0 |
| 9 | Unknown if injured | 51 | 16 | 67 |
| 10 | No person in crash | 1 | 0 | 1 |
| 11 | No PAR (police accident report) obtained | 231 | 50 | 281 |
| **total** | | **3’935** | **3’014** | **6’949** |

*Table 14: Distribution of INJSEVA and INJSEVB*

The length of the case descriptions correlates with the number of vehicles involved, see Figure 29.

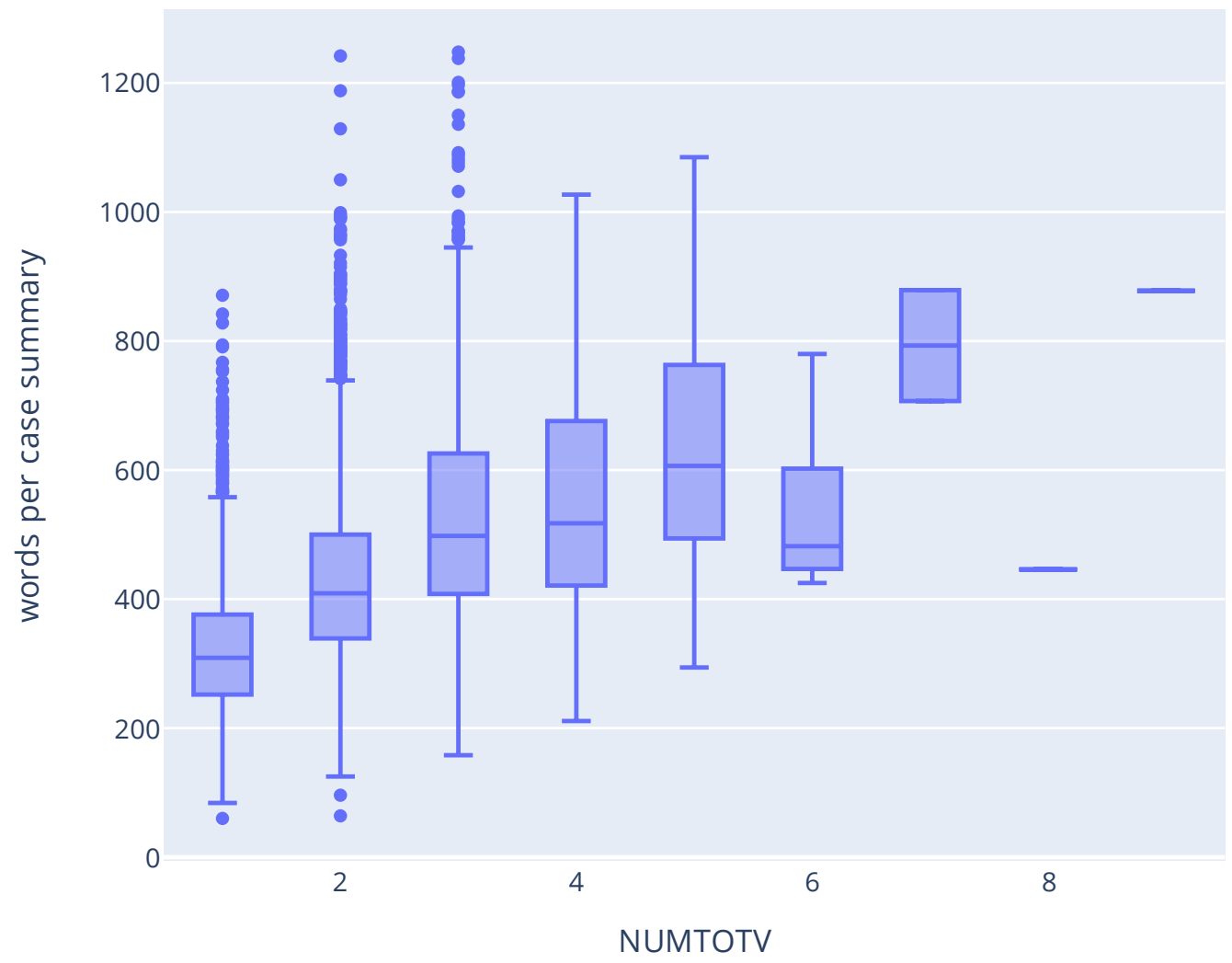

*Figure 29: Box plot of the length of the English case descriptions (counting words split by empty space) by number of vehicles involved. The minimum / average / maximum length is 60 / 419 / 1248 words, respectively.*

Figure 30 shows a sample text, and Figure 31 to Figure 33 show word importance visualizations obtained using the approach described in Section 8.3.

V1, a 2000 Pontiac Montana minivan, made a left turn from a private driveway onto a northbound 5-lane two-way, dry asphalt roadway on a downhill grade. The posted speed limit on this roadway was 80 kmph (50 MPH). V1 entered the roadway by crossing over the two southbound lanes and then entering the third northbound lane, which was a left turn-only lane at a 4-way intersection. The driver of V1 intended to travel straight through the intersection, and so he began to change lanes to the right. He did not see V2, a 1994 Pontiac Grand Am, that was traveling in the second northbound lane. The northbound roadway had curved to the right prior to the private driveway that V1 had exited. As V1 began to change lanes to the right, the front of V1 contacted the left rear of V2 before coming to final rest on the roadway. The driver of V1 was a 60-year old male who reported that he had been traveling between 2-17 kmph (1-10 mph) prior to the crash. He had no health-related problems, and had taken no medication prior to the crash. He was rested and traveling back home. He was wearing his prescribed lenses that corrected a myopic (nearsighted) condition. He did not sustain any injuries from the crash and refused treatment. The Critical Precrash Event for the driver of V1 was when he began to travel over the lane line on the right side of the travel lane. The Critical Reason for the Critical Precrash Event was inadequate surveillance (failed to look, looked but did not see). Associated factors coded to the driver of V1 include an illegal use of a left turn lane (cited by police) and an unfamiliarity with the roadway. As per the driver of V1, this was the first time he had driven on this roadway. The driver of V2 was a 28-year old woman who reported that she had been traveling between 66-80 kmph (41-50 mph) prior to the crash. She had no health-related problems, and had taken no medication prior to the crash. She was rested and on her way home. She does not wear corrective lenses. She sustained minor injuries and was transported to a local trauma facility. The Critical Precrash Event for the driver of V2 was when the other vehicle encroached into her lane, from an adjacent lane (same direction) over the left lane line. The Critical Reason for the Critical Precrash Event was not coded to the driver of V2 and no associated factors were coded to her.

====

V1, ein Minivan der Marke Pontiac Montana aus dem Jahr 2000, bog von einer privaten Einfahrt nach links auf eine zweispurige, trockene Asphaltstraße mit 5 Fahrspuren in nördlicher Richtung und einem Gefälle ab. Die zulässige Höchstgeschwindigkeit auf dieser Fahrbahn betrug 80 km/h (50 MPH). V1 fuhr auf die Fahrbahn, indem er die beiden Fahrspuren in Richtung Süden überquerte und dann auf die dritte Fahrspur in Richtung Norden einfuhr, die an einer Kreuzung mit vier Fahrspuren nur für Linksabbieger vorgesehen war. Der Fahrer von V1 beabsichtigte, geradeaus über die Kreuzung zu fahren, und begann daher, die Spur nach rechts zu wechseln. Dabei übersah er V2, einen Pontiac Grand Am von 1994, der auf der zweiten Fahrspur in Richtung Norden unterwegs war. Die Fahrbahn in nördlicher Richtung war vor der privaten Einfahrt, aus der V1 herausgefahren war, nach rechts gebogen. Als V1 begann, die Spur nach rechts zu wechseln, berührte die Front von V1 das linke Heck von V2, bevor er auf der Fahrbahn zum Stehen kam. Der Fahrer von V1 war ein 60-jähriger Mann, der angab, vor dem Unfall mit einer Geschwindigkeit von 2 bis 17 km/h unterwegs gewesen zu sein. Er hatte keine gesundheitlichen Probleme und hatte vor dem Unfall keine Medikamente eingenommen. Er war ausgeruht und auf dem Weg nach Hause. Er trug die ihm verschriebenen Kontaktlinsen, die eine Kurzsichtigkeit korrigieren. Er zog sich bei dem Unfall keine Verletzungen zu und lehnte eine Behandlung ab. Das kritische Ereignis vor dem Unfall war für den Fahrer von V1, als er begann, die Fahrspurlinie auf der rechten Seite der Fahrbahn zu überfahren. Der kritische Grund für das kritische Ereignis vor dem Unfall war unzureichende Überwachung (nicht hingesehen, hingesehen, aber nicht gesehen). Zu den assoziierten Faktoren, die dem Fahrer von V1 zugeschrieben werden, gehören das illegale Benutzen einer Linksabbiegerspur (von der Polizei verwarnt) und die Unkenntnis der Fahrbahn. Für den Fahrer von V1 war es das erste Mal, dass er diese Fahrbahn befuhr. Bei der Fahrerin von V2 handelte es sich um eine 28-jährige Frau, die angab, vor dem Unfall mit einer Geschwindigkeit von 66-80 km/h unterwegs gewesen zu sein. Sie hatte keine gesundheitlichen Probleme und hatte vor dem Unfall keine Medikamente eingenommen. Sie war ausgeruht und befand sich auf dem Heimweg. Sie trägt keine Korrekturgläser. Sie erlitt leichte Verletzungen und wurde in eine örtliche Unfallklinik gebracht. Das kritische Ereignis vor dem Unfall war für die Fahrerin von V2, als das andere Fahrzeug von einer benachbarten Fahrspur (gleiche Richtung) über die linke Fahrspurlinie in ihre Spur eindrang. Der kritische Grund für das kritische Vorunfallereignis wurde der Fahrerin von V2 nicht zugeordnet, und es wurden ihr keine zugehörigen Faktoren zugeordnet.

*Figure 30: Sample of SUMMARY_EN and SUMMARY_DE (SCASEID= 200501269400).*

**Legend:** Negative, Neutral, Positive

| True Label | Predicted Label | Attribution Label | Attribution Score | Word Importance |
|---|---|---|---|---|
| 1 | LABEL_1 (0.99) | LABEL_1 | 0.87 | [CLS] This crash occurred in the south ##bound lane of a two - lane und ##ivi ##ded road ##way . This was a level asp ##halt road that curve ##d slightly to the left , with a posted speed limit of 64 km ##ph ( 40 mph ) . It was early in the evening on a week ##day , conditions were clear , and the road ##way was dry . There were no traffic flow restrictions . V ##1 was a 2002 Chrysler Se ##bring 2 - door convert ##ible . The vehicle was traveling south ##bound and its driver was beginning to nego ##tia ##te a left curve . V ##1 departed the road ##way to the right and struck a telephone pole located on the roads ##ide . V ##1 rota ##ted clock ##wise after the impact and then trip ##ped over its wheels . V ##1 rolle ##d two quarter - turns and came to final rest on its roof . V ##1 was driven by a 69 - year old female who suffered moderate injuries . The driver has since been put into a nur ##sing home and does not reca ##ll any information from the accident . The accident report and medical records indicated that the driver of V ##1 had a blood alcohol content of 0 . 177 . The Critical Pre - crash Event for V ##1 was this vehicle traveling off the edge of the road on the right side . The Critical Reason for the Critical Pre - crash Event was poor direction ##al control , a driver - related factor . Associated factors code ##d to the driver of V ##1 include alcohol use , the medical condition of diabetes and the use of pre ##scription med ##ication to control the diabetes . Medical reports also indicated that the driver of V ##1 had a history of alcohol ##ism . [SEP] [PAD] [PAD] [PAD] [PAD] [PAD] [PAD] [PAD] [PAD] [PAD] [PAD] [PAD] [PAD] [PAD] [PAD] [PAD] [PAD] [PAD] [PAD] [PAD] [PAD] [PAD] [PAD] [PAD] [PAD] [PAD] |

*Figure 31: Word importance visualization for a true positive example (SCASEID=2006008500862). Apparently, the most important word is "injuries". The original text is padded to a length of 512 tokens (not all shown in the exhibit).*

**Legend:** Negative Neutral Positive

| True Label | Predicted Label | Attribution Label | Attribution Score | Word Importance |
|---|---|---|---|---|
| 1 | **LABEL_1 (0.99)** | **LABEL_1** | **1.32** | [CLS] This three - vehicle crash occurred in the morning of a weekend on a multi - lane highway near an entrance ra ##mp . The highway runs east and west and divided by a high - tension cable guard ##rail . The bit ##umi ##nou ##s road ##way is dry , level and curve ##d to the left at the location of this crash . The posted speed limit 89 km ##ph ( 65 mph ) and there were no ad ##verse weather conditions . V ##1 , a 2006 Je ##ep Liberty with two occupa ##nts , was west ##bound in lane three inte ##nding to go straight . V ##2 , a 1992 Mitsubishi Dia ##mante with one occupa ##nt , was west ##bound in lane four inte ##nding to go straight . V ##3 , a 1996 Nissan pick ##up with one occupa ##nt , was west ##bound in lane one ( ac ##cel ##eration ra ##mp ) inte ##nding to merge left . An unknown vehicle traveling behind V ##3 switched lane ##s and cut in front of V ##1 . V ##1 attempted to avoid this unknown vehicle by changing lane ##s and striking V ##2 ( event # 1 ) . Subsequently , V ##1 and V ##2 sp ##un across all travel lane ##s and departed the right side of the road . V ##1 was struck in the right side by V ##3 as it sp ##un across the ac ##cel ##eration lane and came to final rest on the right roads ##ide . After V ##2 entered the right roads ##ide it sp ##un into an em ##bank ##ment and rolle ##d ( est . 6 - quarter turns ) and came to final rest on its roof . V ##3 drove off the right side of the road after striking V ##1 . The driver of V ##1 is a 45 - year - old female that refused to be interviewed . She was not injured in the crash and her Je ##ep was driven from the scene . The Critical Pre ##cra ##sh Event for V ##1 was code ##d this vehicle traveling over the lane line on the left side of the travel lane . The Critical Reason for the Critical Event was code ##d in ##corre ##ct eva ##sive action . Other factors code ##d to this driver include chose ina ##pp ##rop ##riate eva ##sive action and poor direction ##al control ( failure to control vehicle with skill ord ##inar ##ily expected ) . The driver of V ##2 is a 40 - year - old female that was not interviewed because of a language barrier ( Korean . ) She was transported to the hospital and her vehicle was to ##wed due to damage . The Critical Pre ##cra ##sh Event was code ##d other vehicle en ##cro ##aching from adjacent lane - over right lane line . The Critical Reason for the Critical Event was not code ##d to this vehicle . The driver [SEP] |

*Figure 32: Word importance visualization for a true positive example (SCASEID=2007043731967). The original text is truncated to a length of 512 tokens. Apparently, the most important words are "transported" and "hospital".*

**Legend:** Negative Neutral Positive

| True Label | Predicted Label | Attribution Label | Attribution Score | Word Importance |
|---|---|---|---|---|
| 1 | **LABEL_1 (0.98)** | **LABEL_1** | **0.88** | [CLS] This crash occurred on a straight level bit ##umi ##nou ##s two lane road ##way that was divided by a painted median . The posted speed limit of 72 km ##ph ( 45 mph ) which reduce ##s to 56 km ##ph ( 35 mph ) 100 meters after the crash site . There is a sign indicating the road ##way narrow ##s . The weather was cloud ##y and the road ##way was partially wet . Traffic flow was normal for that time of day . This crash occurred on a week ##day afternoon . Vehicle 1 , a 2002 Nissan Alt ##ima , was traveling behind Vehicle 2 , a 1991 Chevrolet Lu ##mina , when it drove into the safety zone into the on ##coming traffic lane in order to illegal ##ly pass Vehicle 2 . V ##1 returned to its original lane and impact ##ed with V ##2 ' s front left , with its right rear quarter panel . This sp ##un V ##1 in a clock ##wise position 180 degrees , with V ##1 coming to final rest after impact ##ing an em ##bank ##ment on the right side of the road ##way , with its rear left . Vehicle 1 was to ##wed due to damage . V ##1 came to final rest off the road ##way facing in a northeast ##erly direction . V ##2 came to final rest on the road ##way facing in a south ##erly direction . V ##1 was to ##wed due to damage . V ##2 was to ##wed due to its driver going to the hospital with her baby . Vehicle # 1 , the Nissan Alt ##ima , was driven by a belt ##ed 38 - year - old male who refused to be interviewed . He stated he did not want to be both ##ered " with this sh - t " . The Critical Pre ##cra ##sh Event code ##d to Vehicle 1 was : Other - this vehicle traveling entering the road ##way from the left side of the road ##way . The Critical Reason for the Critical Pre ##cra ##sh Event was code ##d as : driver related factor , aggressive driving behavior . Vehicle # 2 , the Chevrolet , was driven by a belt ##ed 21 year - old female who was not injured . There was a belt ##ed 18 year - old male in the front right seat who was not injured . There was a 6 - month - old female child in a car seat in the second row . The child was taken to the hospital for a check out , accompanied by both other people in the vehicle . This driver stated to her relative that she had seen the driver of V ##1 making " wild ge ##stu ##res " and tail ##gating her . She stated she saw V ##1 coming around her on the left but could only brak ##e before impact . The Critical Pre ##cra ##sh Event code [SEP] |

*Figure 33: Word importance visualization for a false positive example (SCASEID= 2007002229650). The original text is truncated to a length of 512 tokens. Apparently, the words "was taken" and "hospital" are most important, as in Figure 32. However, in this case it was only for a check out, and as such there is no strong evidence of a bodily injury in this case description.*

## 13.2 Wisconsin Local Government Property Insurance Fund

This dataset concerns property insurance claims of the Wisconsin Local Government Property Insurance Fund (LGPIF), made available by [Frees2020][24]. The Wisconsin LGPIF is an insurance pool managed by the Wisconsin Office of the Insurance Commissioner. This fund provides insurance protection to local governmental institutions such as counties, schools, libraries, airports, etc. It insures property claims at buildings and motor vehicles, and it excludes certain natural and manmade perils like flood, earthquakes or nuclear accidents.

The data consists of 6'030 records (4'991 in the training set, 1'039 in the test set) which include a claim amount, a short English claim description and a hazard type with 9 different levels: Fire, Lightning, Hail, Wind, WaterW (weather related water claims), WaterNW (other weather claims), Vehicle, Vandalism and Misc (any other). The following exhibit shows an example.

| row | Vandalism | Fire | Lightning | Wind | Hail | Vehicle | WaterNW | WaterW | Misc | Loss | Description |
|---|---|---|---|---|---|---|---|---|---|---|---|
| 1 | 0 | 0 | 1 | 0 | 0 | 0 | 0 | 0 | 0 | 6838.87 | lightning damage |
| 2 | 0 | 0 | 1 | 0 | 0 | 0 | 0 | 0 | 0 | 2085 | lightning damage at Comm. Center |
| 6 | 1 | 0 | 0 | 0 | 0 | 0 | 0 | 0 | 0 | 8775 | surveillance equipment stolen |
| 7 | 0 | 0 | 0 | 1 | 0 | 0 | 0 | 0 | 0 | 34610.27 | wind blew stack off and damaged roof |
| 9 | 0 | 0 | 0 | 0 | 0 | 1 | 0 | 0 | 0 | 9711.28 | forklift hit building damaging wall and door frame |
| 11 | 0 | 0 | 0 | 0 | 0 | 0 | 0 | 1 | 0 | 1942.67 | water damage at courthouse |
| 30 | 0 | 0 | 0 | 0 | 0 | 1 | 0 | 0 | 0 | 3469.79 | light pole damaged |

*Table 15: Sample of the LPGIF data set*

For the examples shown, the peril classification is plausible, given the text description. The exception is row 11 which could be attributed to WaterNW as well.

[24] https://github.com/OpenActTexts/Loss-Data-Analytics/tree/master/Data